\documentclass[10pt,journal,compsoc]{IEEEtran}

\usepackage{changepage}
\usepackage{amsfonts} 
\usepackage{amsmath} 
\usepackage{graphicx}
\usepackage{multirow, tabularx}
\usepackage{color}
\usepackage[]{algorithm2e}
\usepackage{threeparttable}
\usepackage{url}
\usepackage{slashbox}
\usepackage[most]{tcolorbox}
\usepackage{amsmath}

\usepackage[table]{colortbl}

\usepackage{caption}
\usepackage{subcaption}

\usepackage[nomessages]{fp}

\newcommand{\colorMyCellMEDS}[1] {
\FPeval{\colorValue}{(#1*100-0)/(0.162-0)}
\xdef\colorValue{\colorValue}
\cellcolor{red!\colorValue}
\scriptsize#1
}

\newcommand{\colorMyCellMORPH}[1] {
\FPeval{\colorValue}{(#1*100-0)/(0.314-0)}
\xdef\colorValue{\colorValue}
\cellcolor{red!\colorValue}
\scriptsize#1
}

\newcommand{\colorMyCellMOBIO}[1] {
\FPeval{\colorValue}{(#1*100-0)/(0.28-0)}
\xdef\colorValue{\colorValue}
\cellcolor{red!\colorValue}
\scriptsize#1
}

\newcommand{\colorMyCell}[1] {
\FPeval{\colorValue}{#1*100}
\xdef\colorValue{\colorValue}
\cellcolor{red!\colorValue}
\scriptsize#1
}

\ifCLASSOPTIONcompsoc
  \usepackage[nocompress]{cite}
\else
  \usepackage{cite}
\fi

\newtcolorbox{snfadvice}{%
    title=Note,%
    before skip=0.5em,%
    after skip=2em,%
    fonttitle=\sffamily\bfseries\large,%
    sharp corners,%
    colframe=gray!75!white,%
    parbox=false}

\begin{document}

\markboth{}{}

\title{Fairness in Biometrics:  a figure of merit to assess biometric verification systems}

\author{Tiago de Freitas Pereira, \and S{\'e}bastien Marcel\\
        Idiap Research Institute \\
        \small{\{tiago.pereira,sebastien.marcel\}@idiap.ch}



}


\IEEEtitleabstractindextext{%
\begin{abstract}

Machine learning-based (ML) systems are being largely deployed since the last decade in a myriad of scenarios impacting several instances in our daily lives.
With this vast sort of applications, aspects of fairness start to rise in the spotlight due to the social impact that this can get in some social groups.
In this work aspects of fairness in biometrics are addressed.
First, we introduce the first figure of merit that is able to evaluate and compare fairness aspects between multiple biometric verification systems, the so-called Fairness Discrepancy Rate (FDR).
A use case with two synthetic biometric systems is introduced and demonstrates the potential of this figure of merit in extreme cases of demographic differentials.
Second, a use case using face biometrics is presented where several systems are evaluated compared with this new figure of merit using three public datasets exploring gender and race demographics.

\end{abstract}

\begin{IEEEkeywords}
Biometrics, Fairness, Face Recognition
\end{IEEEkeywords}}

\newtheorem{premisse}{Premise}

\maketitle

\IEEEdisplaynontitleabstractindextext

\IEEEpeerreviewmaketitle

\IEEEraisesectionheading{\section{Introduction}\label{sec:introduction}}


\IEEEPARstart{T}he pipeline from research to deployment of an ML-based system can assume several shapes with different steps.
In abstract terms (and allow us to do such simplification), such pipeline is composed of  i-) \textbf{Data Collection}: where the ``state of the world'' is reduced to a set of rows and columns of data (e.g. face images, bank transactions, medical data, etc...); ii-) \textbf{Modelling}: where the ``model'' is supposed to summarize the patterns of the data and be able to make generalizations (via supervised/unsupervised learning, etc..); iii-) \textbf{Benchmarking}: where the model is evaluated with respect to some figure of merit (e.g. accuracy, f1-score, etc..); iv-): \textbf{Feedback} where it is decided if the model is ``good'' for deployment or not; if not, steps (i) and/or (ii) needs to be redone; v-) \textbf{Deployment}: ML-System goes to production\footnote{Usually, feedbacks are also done after deployment, but let's keep this simplification as is because it is enough for our purposes.}.
During the benchmarking stage, it is common to use reference databases.
Such reference databases are supposed to represent somehow operational conditions and it is hypothesized that ML-based systems that presents high accuracy, high f1 score, low false-positive rate, low false-negative rate, etc in such benchmarks is a proxy to have the same figures of merit in operational conditions.
Once this is achieved (by any criteria ML engineers decide), ML is ``safe'' to be deployed.

Fairness issues arise from the analysis of these figures of merit in specific demographics groups (e.g., gender, ethnicity, race, revenue levels, or any covariate in general) and the observation that operational conditions estimated initially cannot be reproduced in those.
In biometrics, this is coined as demographic differentials.
The large-scale deployment of such systems in so many different scenarios raises the debate about how these differentials impact people's lives.

For instance, the book Weapons of Math Destruction \cite{o2016weapons} presents several cases where unfair decision-making tools based on ML impacted the life of city populations negatively.
Such cases cover different applications, such as online advertising tools, automatic resum{\'e} evaluations for HR, and credit score tools for different purposes (e.g., bank or insurance companies).

\begin{figure}
  \begin{subfigure}[b]{0.4\columnwidth}
    \includegraphics[width=\linewidth]{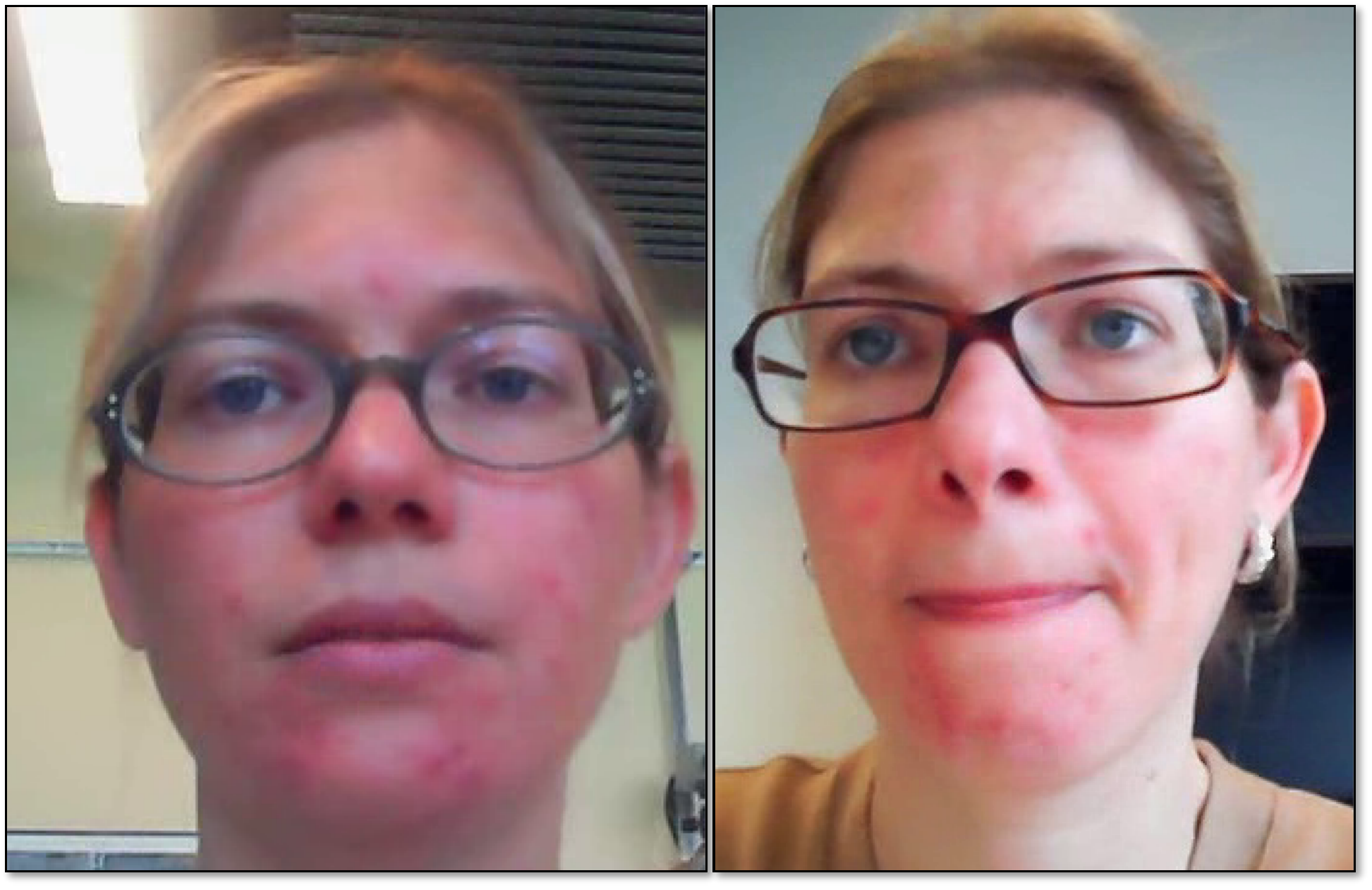}
    \caption{Score: \textbf{\color{red}{-0.6445}}}
    \label{fig:1}
  \end{subfigure}
  \hfill 
  \begin{subfigure}[b]{0.4\columnwidth}
    \includegraphics[width=\linewidth]{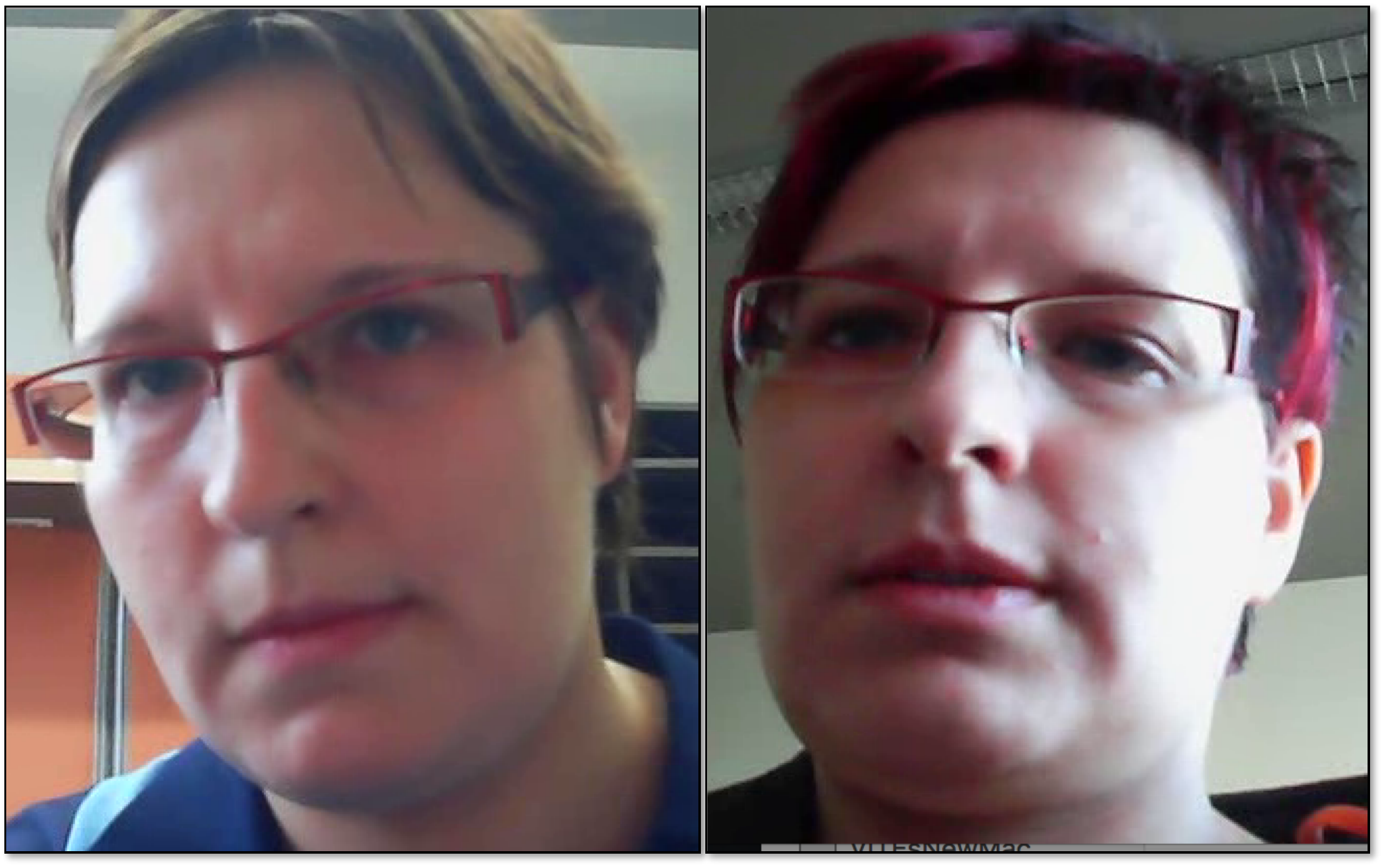}
    \caption{Score: \textbf{\color{red}{-0.6307}}}
    \label{fig:2}
  \end{subfigure}

  \begin{subfigure}[b]{0.4\columnwidth}
    \includegraphics[width=\linewidth]{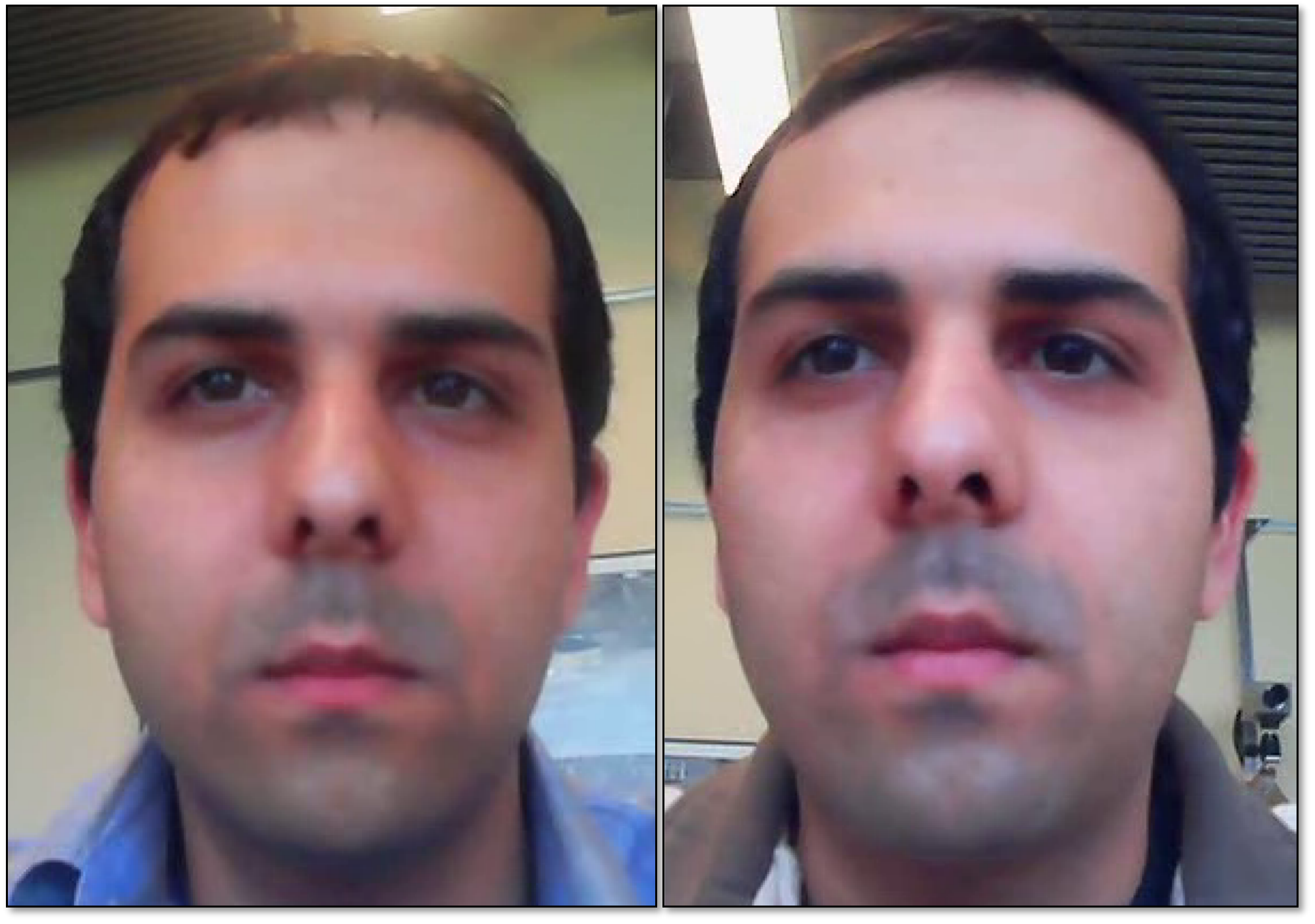}
    \caption{Score: \textbf{\color{blue}{-0.1596}}}
    \label{fig:3}
  \end{subfigure}
  \hfill 
  \begin{subfigure}[b]{0.4\columnwidth}
    \includegraphics[width=\linewidth]{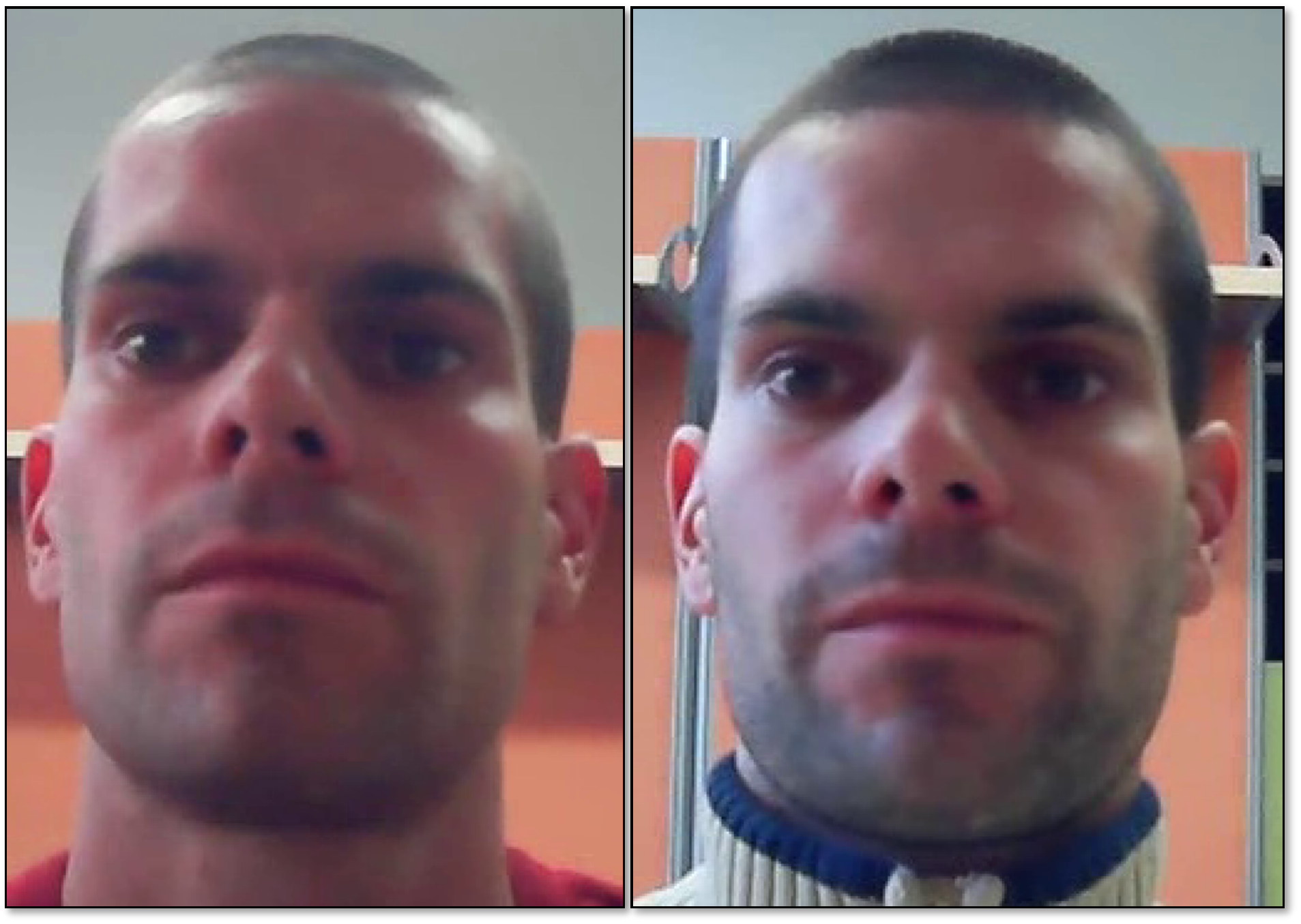}
    \caption{Score: \textbf{\color{blue}{-0.1679}}}
    \label{fig:4}
  \end{subfigure}
  \caption{Inception Resnet v2 FR system\cite{szegedy2016inception}: Genuine comparison pairs extracted MOBIO Database\cite{mccool2012bi}. The decision threshold is: $\tau=-0.5298$. Hence, $\text{score} \geq \tau$, the comparison pair is accepted; otherwise it is rejected.}
  \label{fig:fairness_mobio_example}
\end{figure}

Decision-making tools based on biometrics, as part of this Machine Learning wave, have been largely deployed in the recent decade.
For instance, it is present in our daily lives for data protection (e.g., unlock mobile phones or computers), law enforcement, airport e-gates, and other applications.
The public media has been reported several cases where biometric systems, specially Face Recognition (FR) technology, present recognition disparaties between different demographic groups.
For instance, a simple test executed in 2018 using a Commercial-Off-the-Shelf (COTS) system comparing face images from members of the US parliament found 28 false matches, and most of them occur with ``people of color''\footnote{https://www.aclu.org/blog/privacy-technology/surveillance-technologies/amazons-face-recognition-falsely-matched-28}.
In 2020, the Association for Computing Machinery in New York City recommended the suspension of private and government use of FR technology due to ``clear bias based on ethnic, racial, gender, and other human characteristics''\footnote{https://www.acm.org/binaries/content/assets/public-policy/ustpc-facial-recognition-tech-statement.pdf}.

This work addresses demographic differential aspects in biometric systems, and its contributions are twofold.
First, it discusses the factors to consider a biometric verification system as fair and introduces the first figure of merit in this field, the Fairness Discrepancy Rate.
Second, a case study of this figure of merit is presented using face recognition as a biometric trait.
We aim to make this reproducible: all the source code, trained models, and scores are made publicly available.
Details on how to reproduce this work can be found on the provided link\footnote{https://gitlab.idiap.ch/bob/bob.paper.fdr}.

\section{Related work}
\label{sec:related_work}

In this section, we present the related work by first discussing the Machine Learning community's efforts to suppress demographic differentials, and then we move to actions made by the biometrics community in this direction.

\subsection{Machine Learning Background}

Many criteria to assess and address differentials in pattern recognition problems have been proposed over the years, each one phrasing the problem in different ways.
The recent work from \cite{barocas-hardt-narayanan} hypothesizes that most of these criteria described in the machine learning literature falls into three major categories of conditional independence and they are: Independence, Separation, and Sufficiency.

To illustrate these criteria, let's consider $X \in \mathbb{R}^n$ a random variable denoting the input data, $D=\{d_1, d_2 ... d_n\}$ a random variable denoting a set of sensitive attributes (e.g. gender, demographics, etc), $Y \in \{0,1\}$ (for simplicity) a random variable denoting the target variable (representing a binary classifier) and $F: f(X,D)$ the trained predictor (that can be possibly thresholded).
The first non-discrimination criteria, and the most simplistic one, is \textbf{independence} which requires that the classifier $F$ must be independent of the sensitive attributes $D$, or $F \perp D$.
This is also addressed as \textit{demographic parity} or \textit{statistical parity}.
For our binary classification case, this can be rewritten as:

\begin{equation}
P\{F=1 | D=d_i \} =  P\{F=1 | D=d_j \} \forall \{i,j\} \in D.
\end{equation}

This criterion is largely used in ML, in general, to mitigate differentials either via regularization \cite{wang2019racial,chen2020towards}, representation learning constraints \cite{louizos2015variational,locatello2019fairness,zemel2013learning}, or post-processing mechanisms \cite{feldman2015certifying}.
The indepence criteria, although popular, it has some limitations.
For instance, at training-time, a classifier might trade false positives to false negatives in a group to match independence.
Other limitations of independence are largely discussed in \cite{hardt2016equality} and more recently in \cite{barocas-hardt-narayanan}.

The second criterion is \textbf{separation} (or equalized odds criteria) where it explicitly acknowledges that the target variable $Y$ might be correlated with $D$.
Conditioning in $Y$ might be desirable in some scenarios.
For instance, a medical doctor might argue that a particular disease is more likely to be developed in one demographic group than others, and a ``disease'' prediction function $F$ must take this into account.
The following condition independence summarizes this: $F \perp D | Y$.
For our binary classification case, this is equivalent to:

\begin{equation}
   \begin{aligned}
      & P\{F=1 | Y=y,D=d_i\} = P\{F=1 | Y=y,D=d_j \} \\
      & \forall \{i,j\} \in D \\
      & y \in \{0, 1\} 
   \end{aligned}
   \label{eq:sep_1}
\end{equation}

What separation requires is that all demographic groups should experience the same true/false-positive rates \cite{barocas-hardt-narayanan}[sec 2, p. 12].
The work from \cite{hardt2016equality} studied different formulations of separation (e.g., Equalized Odds and Equality of Opportunity) and proposed different constrained optimization approaches to reach separation.
Using FICO scores as a case study demonstrates that such criterion can ``remediate'' some of the problems from methods based on independence.
The authors from \cite{zafar2017fairness} followed a similar path.
In \cite{chen2020towards} the authors propose different regularization strategies to reach independence by making risk scores threshold independent.

The third criterion is \textbf{sufficiency}.
$F$ is sufficient to predict $Y$ independently of the demographic attribute $D$ if the following conditional probability is matched: $Y \perp D | F$.

In this case:

\begin{equation}
   \begin{aligned}
      & P\{Y=1 | F=s,D=d_i\} =  P\{Y=1 | F=s,D=d_j\}\\
      & \forall \{i,j\} \in D.
   \end{aligned}
\end{equation}

Most of the literature that approaches sufficiency estimators does so via calibration of the risk scores by group.
One of the most popular methods to perform such calibration is Platt scaling \cite{platt1999probabilistic}, which, roughly, consists of fitting a sigmoid function on uncalibrated scores.
More recently, in \cite{pleiss2017fairness} the authors propose different optimization mechanisms to achieve sufficiency.

Those three essential fairness criteria support most of what was published in the Machine Learning literature explicitly or implicitly.

\subsection{Fairness in Biometrics}

In the biometrics literature, some recent work addresses demographic differentials for some biometric traits.
For instance, the Face Recognition Vendor Test (FRVT) has a special report addressing demographic effects in FR \cite{grother2019face}.
This report presents several analyses observing the effect, mostly of race and gender in terms of False Matches and False non Matches with more than 100 COTS FR systems.
More recently, a comprehensive survey on demographics differentials on biometrics \cite{drozdowski2020demographic} demonstrated efforts in this direction on the face, fingerprint, palm vein, iris, and palm print recognition research.

The recent work from \cite{cavazos2019accuracy} describes some underlying factors that biases COTS face recognition systems concerning \textbf{race}.
For instance, they observed that the ``Other Race Effect'', well known in humans, \cite{malpass1969recognition} can also be observed in FR algorithms; FR systems developed in Asia are more accurate with Asian subjects than with Caucasians, and vice-versa.
Furthermore, they observed that demographic differentials are frequently observed where low-quality probe samples are used.
The FRVT report also raised such observations about image quality.
Studying race, the work from \cite{vangara2019characterizing} observed consistently higher False Match Rates (FMR) with African American cohorts compared with Caucasians using two COTS systems.
Furthermore, this work extended its analysis with ICAO face checker\footnote{https://www.icao.int/Security/FAL/TRIP/Documents/TR - Portrait Quality v1.0.pdf}.
They observed that ICAO SDKs work better with Caucasian subjects than with African Americans.
The work from \cite{michalski2018impact} made an extensive study analyzing several \textbf{age} cohorts using one COTS system.
Among several observations made, the most impacting one was the high FMR and high False Non-Match Rates (FNMR) in pairs of images where age is lower than four years old.

\begin{figure}[t]
     \begin{center}

        \includegraphics [width=26em] {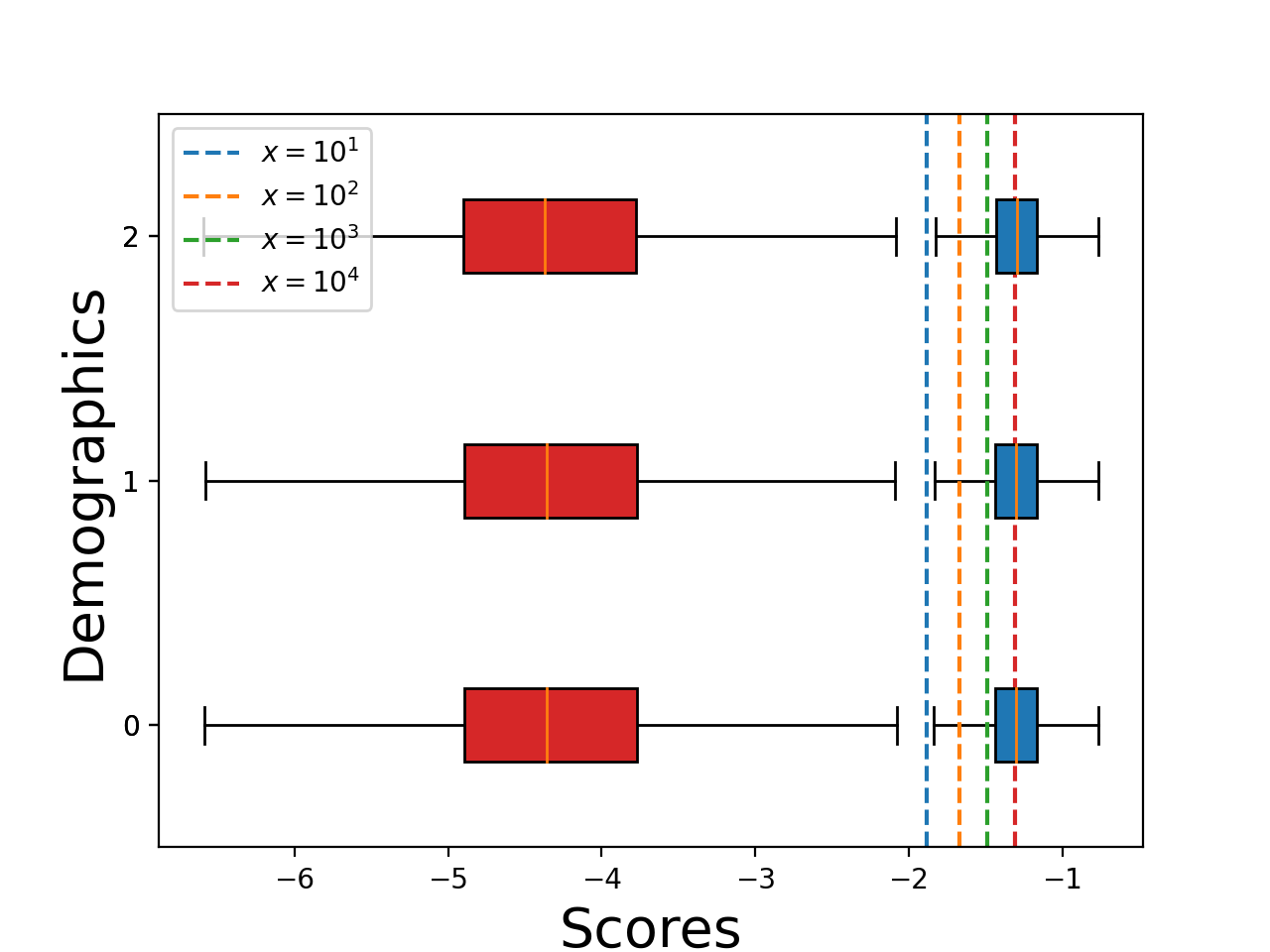}
        \caption{Example of a canonical fair biometric verification system with three demographics (0, 1, 2) and six operational thresholds (depicted with the dashed lines). Performance measures in terms of $\text{FMR}(\tau)$ and $\text{FNMR}(\tau)$ can be found in TABLE \ref{tb:fair}.}
        \label{fig:fair_abstract_box_plot}
     \end{center}
\end{figure}

Focusing on face biometrics, the work from \cite{wang2019racial} introduces the Racial Faces in the Wild dataset.
Such dataset is a subset of the MSCeleb-1M \cite{guo2016ms} whose identities are organized in four different races (Caucasians, Black, Indian and Chinese).
Using such data, the authors, at training-time, regularized different deep neural networks by minimizing the Mutual Information between the face recognition classifier and the demographic attributes. 
In \cite{kortylewski2019analyzing} the author's leverage from synthetically generated data to train fairer DCNNs for FR.
The authors from \cite{amini2019uncovering} propose a framework to automatically calibrate the weight of samples at training-time of a DCNN based on protected attributes.
In \cite{rudd2016moon} the authors followed a similar path in the task of facial attributes classification.
The authors from \cite{terhorst2020post} propose a post-processing mechanism based on score normalization to deal with demographic discrepancies.

Demographic differentials towards gender were observed using periocular biometrics.
For instance, the work from \cite{de2015periocular} demonstrates that several periocular recognition systems perform better with male subjects than with female ones.

The NIST SRE\footnote{https://sre.nist.gov/} is the most relevant benchmark for speaker recognition, and along with last editions, it consistently evaluates error rates looking at gender cohorts.

To the best of our knowledge, the works from biometric literature that addresses somehow demographic differentials by either analyzing COTS systems or proposing a strategy to mitigate it do so using different criteria.
However, the trend seems to achieve somehow the \textit{statistical parity} (or independence), even if this detail is not explicitly mentioned.
Even if this is the trend, a figure of merit to directly address it is nonexistent.
For instance, the work from \cite{gong2019debface} uses the Area Under the ROC curve as a figure of merit to measure demographic differentials.
ROC curves measure the True and False Positive Rates (TPR and FPR, respectively) trade-offs.
Although this seems sensible to assess demographic discrepancies, it has a serious flaw; it assumes that the verification decision threshold (let us call it $\tau$; we will formally define this further) is demographic-specific.
Hence, $\text{TPR}(\tau)$ and $\text{FPR}(\tau)$ is computed under different decision thresholds depending of the demographic and can give a false impression that a biometric verification system is fair (this problem is further discussed in section \ref{sec:fdr_chapter}).
Furthermore, this does not represent operational conditions where normally one single $\tau$ is set, and this operational point has to be fair for different demographics.
This issue with ROC curves was recently raised in \cite{krishnapriya2020issues}.
This problem can be observed also in several works that refers to biometric verification; for instance, in \cite{de2015periocular,robinson2020face,wang2019racial,wang2020mitigating,cavazos2019accuracy}.

\begin{table}[t]
   \begin{center}

   \caption[]{Canonical fair biometric verification system: $\text{FNMR}(\tau)$, $\text{FMR}(\tau)$, and $\text{FDR}(\tau)$ per demographic (Demog.) where the operational points are defined as $\tau=\text{FMR}_{x}$\textsuperscript{*}}
    \begin{tabular}{|r||rrrr|}
      \hline
  
      $\tau=FMR_{10^{x}}$ &  $10^{-3}$ & $10^{-4}$ & $10^{-5}$ & $10^{-6}$\\
	
      \hline
      Demog. & \multicolumn{4}{c|}{$\text{FMR}(\tau)$} \\
      \hline

  0 &  \colorMyCell{0.001}  &   \colorMyCell{0.000} &       \colorMyCell{0.000}  &       \colorMyCell{0.000}  \\
  1 &  \colorMyCell{0.001}  &   \colorMyCell{0.000} &       \colorMyCell{0.000}  &       \colorMyCell{0.000}  \\
  2 &  \colorMyCell{0.001} &   \colorMyCell{0.000} &       \colorMyCell{0.000}  &       \colorMyCell{0.000}  \\

      \hline
      & \multicolumn{4}{c|}{$\text{FNMR}(\tau)$} \\

      \hline
  0 &     \colorMyCell{0.003} &   \colorMyCell{0.036} &  \colorMyCell{0.179} &  \colorMyCell{0.484} \\
  1 &     \colorMyCell{0.003} &   \colorMyCell{0.039} &  \colorMyCell{0.178} &  \colorMyCell{0.481} \\
  2 &     \colorMyCell{0.003} &   \colorMyCell{0.036} &  \colorMyCell{0.168} &  \colorMyCell{0.469} \\
       \hline
        $FDR(\tau)$ &  0.999 &   0.999 & 0.994 &  0.992 \\
       \hline
      
   \end{tabular}
  \label{tb:fair}
  \end{center}
  \small\textsuperscript{\textsuperscript{*}$\tau$ is set using an independent zeroth-effort impostor score distribution}
  \small\textsuperscript{with scores from all demographics. It can be seen as a development set.}  
\end{table}

Some works in the biometrics literature \textbf{explicitly} analyses the possibility of $\tau$ being demographic-specific, such as in \cite{cook2019demographic,poh2010group}.
Even in \cite[(sec.2, p.14)]{barocas-hardt-narayanan} (covering a general case of pattern recognition), the authors work with the possibility of one $\tau$ per demographic.
For biometric recognition, strategies in this direction might not be ethical, since it imposes a disparate treatment on deployed systems.
Furthermore, in practical terms, this can involve another classification task in the biometric recognition pipeline, which might be error-prone and subject to disparities.

FRVT goes in the right direction concerning the aforementioned threshold problem by discussing the impact of demographics in terms of $\text{FMR}(\tau)$ and $\text{FNMR}(\tau)$ for one decision threshold only.
Such a decision threshold is picked from an independent zero-effort score distribution, where the demographic does not play a role.
This is the most sensible evaluation if the goal is to assess demographic differentials in operational conditions.
However, FRVT discussess the impact of $\text{FMR}(\tau)$ and $\text{FNMR}(\tau)$ \textbf{separately}.
Hence, the trade-off between them is not considered.
Furthermore, only one decision threshold is analyzed, limiting the perception of demographic differentials under different operational points.
In \cite{chen2020towards} a similar direction was taken where risk distributions among the different demographic groups were equalized via different approximation methods, introducing then threshold invariant classifiers.
However, no analysis in terms $\text{FMR}(\tau)$ and $\text{FNMR}(\tau)$ was carried out.

Our work tries to fill these evaluation gaps for biometric verification systems by: (i) - taking into consideration the above-mentioned threshold problems (ii) - considering the $\text{FMR}(\tau)$ and $\text{FNMR}(\tau)$ trade-off in the demographic differential assessment, and (iii) - taking into account different operation points (decision thresholds).

\begin{figure}[t]
     \begin{center}

        \includegraphics [width=26em] {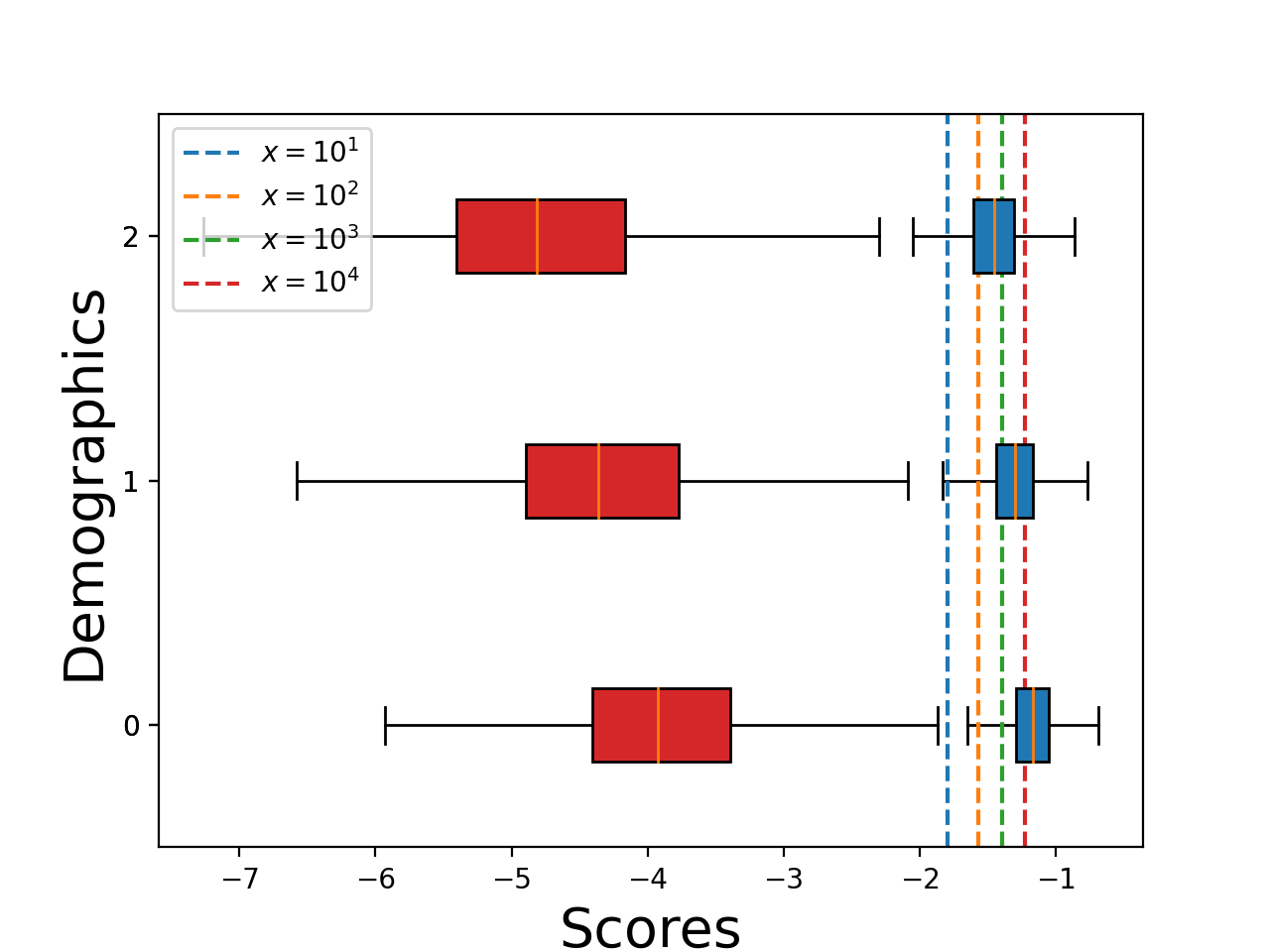}
        \caption{Example of a canonical UNfair biometric verification system with three demographics (0, 1, 2) and six operational  points (depicted with the dashed lines). Performance measures in terms of $\text{FMR}(\tau)$ and $\text{FNMR}(\tau)$ can be found in TABLE \ref{tb:unfair}.}
        \label{fig:unfair_abstract_box_plot}
     \end{center}
     \vspace{-1em}
\end{figure}

\section{Fairness Discrepancy Rate: A figure of merit to assess fairness in biometric verification}
\label{sec:fdr_chapter}

Biometric verification is the task of verifying if a given sample is from a claimed identity or not.
This decision is made based on a scoring function $s(e,p)$ and a decision threshold $\tau$, where $e$ is the claimed identity, $p$ is a probe sample (test sample).
If $s(e,p) \geq \tau$ it is said that $e$ and $p$ are from the same identity.
Conversely, if $s(e,p) < \tau$ it is said that $e$ and $p$ are not from the same identity.
There are two possible types of errors that biometric verification systems can make, and they are the False Match Rate (\textbf{FMR}) and False Non-Match Rate (\textbf{FNMR}).
Worth noting that these two errors are functions of a decision threshold $\tau$, whose impact is further discussed.

The value of $\tau$ plays a decisive role in these two errors, and it is usually set targeting a specific $\text{FMR}$ value in a reference impostor score distribution set\footnote{Impostor score distribution is made of $s(e,p)$ values where $e$ and $p$ are not from the same identity}.
Some examples of such operational points are: $\tau=\text{FMR}_{10^{-1}}$ corresponds to the $\tau$ where FMR reaches $0.1$ (or 10\%) in the impostor distribution scores; $\tau=\text{FMR}_{10^{-3}}$ corresponds to the $\tau$ where FMR reaches $0.001$ (or 0.1\%) in the impostor distribution scores; $\tau=\text{FMR}_{10^{-6}}$ corresponds to the $\tau$ where FMR reaches $10^{-6}$ (or 0.001\%) in the impostor distribution scores\footnote{Examples on how this is set in practice can be seen in this report https://pages.nist.gov/frvt/reports/11/frvt\_11\_report.pdf section 1.1}.

Given a test set, ``good'' biometric recognition systems should present $\text{FMR}_{x}(\tau)$ around the operational point given by $x$ and the lowest value as possible for $\text{FNMR}(\tau)$.
Furthermore, for a ``good'' biometric system to be considered fair, it should present $\text{FMR}_{x}(\tau)$ around the operational point $x$ for all observed demographic groups and approximately ``same'' $\text{FNMR}(\tau)$ for all observed demographic groups.
The impact of the decision thresholds is illustrated in Figure \ref{fig:fairness_mobio_example}.
In this example, we chose two comparison scores from male and female subjects of the MOBIO dataset using one of our tested Deep Convolutional Neural Network (DCNN) (see section \ref{sec:fv_usecase} for further details).
Those genuine pairs were cherry-picked by looking at the score values around the average genuine scores for each demographic group.
$\tau$, in this case, equals $-0.5298$.
It can be noticed that both comparisons using female subjects are rejected using this operational point, and the two male subjects are accepted.

\begin{table}[t]
   \begin{center}

   \caption{Canonical UNfair biometric verification system: $\text{FNMR}(\tau)$, $\text{FMR}(\tau)$, and $\text{FDR}(\tau)$ per demographic (Demog.) where the operational points are defined as $\tau=\text{FMR}_{x}$\textsuperscript{*}}

    \begin{tabular}{|r||rrrr|}
      \hline
  
     $\tau=FMR_{10^{x}}$ &   $10^{-3}$ & $10^{-4}$ & $10^{-5}$ & $10^{-6}$\\
	
      \hline
      Demog. & \multicolumn{4}{c|}{$\text{FMR}^{\text{test}}(\tau)$} \\
      \hline
  0 &  \colorMyCell{0.002}  &   \colorMyCell{0.002} &       \colorMyCell{0.000} &       \colorMyCell{0.000} \\
  1 &   \colorMyCell{0.000}  &   \colorMyCell{0.000} &       \colorMyCell{0.000} &       \colorMyCell{0.000} \\
  2 &   \colorMyCell{0.000}  &   \colorMyCell{0.000} &       \colorMyCell{0.000} &       \colorMyCell{0.000} \\

      \hline
      & \multicolumn{4}{c|}{$\text{FNMR}^{\text{test}}(\tau)$} \\
      
      \hline
  0 &   \colorMyCell{0.001} &   \colorMyCell{0.016} &  \colorMyCell{0.114} &  \colorMyCell{0.369}   \\
  1 &   \colorMyCell{0.009} &   \colorMyCell{0.092} &  \colorMyCell{0.321} &  \colorMyCell{0.637} \\
  2 &   \colorMyCell{0.071} &   \colorMyCell{0.304} &  \colorMyCell{0.614} &  \colorMyCell{0.855} \\

       \hline
        $FDR(\tau)$ &  0.963 &  0.855 &  0.750 &  0.756 \\
       \hline        
      
   \end{tabular}

  \label{tb:unfair}
  \end{center}
  \small\textsuperscript{\textsuperscript{*}$\tau$ is set using an independent zeroth-effort impostor score distribution}
  \small\textsuperscript{with scores from all demographics. It can be seen as a development set.}      
\end{table}

In this work, a biometric verification system is considered fair if different demographic groups share the same $FMR$ and $FNMR$ for a given decision threshold $\tau$. 
This can be interpreted as an assessment of separation or more specifically to equalized odds \cite{hardt2016equality}.
More formally, given a set of demographic groups $\mathcal{D}=\{d_1,d_2,...,d_n\}$, and $\tau = \text{FMR}_{x}$\footnote{$\tau$ is set using an independent zeroth-effort impostor score distribution with scores from all demographics}, a biometric verification system is considered fair with respect to $\text{FMR}$ if the following premisse holds:

\vspace{1mm}
\begin{premisse}
 \label{pre:A}
  $\text{FMR}^{d_i}(\tau) \geq  \text{FMR}^{d_j}(\tau) - \epsilon \text{ } \forall d_i,d_j \in D$.
\end{premisse}

Such premise can be written with the following equation:
\begin{equation}
 A(\tau) = \max(|\text{FMR}^{d_i}(\tau)- \text{FMR}^{d_j}(\tau)|) \leq \epsilon \text{ } \forall d_i, d_j \in \mathcal{D},
  \label{eq:preA}
\end{equation}
where $\epsilon$ is a relaxation constraint.

Conversely, in terms of $\text{FNMR}$, a biometric verification system is considered fair if the following premise holds:
\vspace{1mm}
\begin{premisse}
 \label{pre:B}
  $\text{FNMR}^{d_i}(\tau) \geq  \text{FNMR}^{d_j}(\tau) \text{ } \forall d_i,d_j \in D$.
\end{premisse}

Such premise can be written with the following equation:
\begin{equation}
 B(\tau) = \max(|\text{FNMR}^{d_i}(\tau)- \text{FNMR}^{d_j}(\tau)|) \leq \epsilon \text{ } \forall d_i, d_j \in \mathcal{D}.
 \label{eq:preB}
\end{equation}

Since \ref{eq:preA} and \ref{eq:preB} are functions of $\tau$, both can be summarized in one figure of merit, that we refer as Fairness Discrepancy Rate (FDR) which is defined as:

\begin{equation}
 FDR(\tau) = 1- (\alpha A(\tau) + (1-\alpha) B(\tau)),
 \label{eq:fairness}
\end{equation}
where $\alpha$ is a hyper-parameter that defines the weight of $A(\tau)$ in the figure of merit (the importance of False Matches).

The values that FDR can take varies from $0$ (maximum discrepancy between two demographic groups) to $1$ (minimum discrepancy between two demographic groups).
In the computation of $A(\tau)$, only false-matches from biometric-references and probe pairs from homogeneous groups are considered (e.g., comparison scores between Black-White, Female-Male, \ldots samples).
Several works in the literature (\cite{grother2019face,howard2019effect,michalski2018impact}) show that the number of false-matches from non-homogenous groups is substantially lower than with homogenous groups.
Hence, to enforce parity between homogenous and non-homogeneous groups seems counter-productive since non-homogeneous comparison pairs present naturally lower false-matches.
As with equations \ref{eq:preA} and \ref{eq:preB} FDR can be possibly thresholded with a slack variable $\epsilon$ and an overall threshold defining what is fair and what is not can be defined as:

\begin{equation}
    \begin{cases}
      \text{fair} & \text{if } \text{FDR}(\tau) \geq 1-\epsilon \\
      \text{unfair} & \text{otherwise}
    \end{cases} 
    \label{eq:equality}
\end{equation}
The role of $\epsilon$ is discussed further in this section.

The following subsection presents one example of a desired fair biometric recognition system and one example of an undesired unfair biometric verification system that illustrates how FDR evaluates these two systems.

\subsection{Fairness Discrepancy Rate using synthetic data}

Figure \ref{fig:fair_abstract_box_plot} shows a canonical fictional example of a fair biometric recognition system.
Each box plot shows the score distributions from both zeroth effort impostors (in red) and genuines (in blue) of three abstract demographics (labeled as 0, 1, and 2).
It is possible to visually inspect that the score distribution from the three demographics is systematically aligned in all quartiles, which indicates that Premisses \ref{pre:A} and \ref{pre:B} can hold for both $\text{FMR}$ and $\text{FNMR}$ for any given $\tau$.
In this experiment $\tau = \text{FMR}_{x}(\tau)$ where $x$ varies from $10^{-3}$ to $10^{-6}$.
Conversely, on the other side of the spectrum, an example of an unfair biometric verification system is presented in Figure \ref{fig:unfair_abstract_box_plot}.
As it can be noticed, the score distributions from both zeroth effort impostors and genuines are not as aligned as in the previous example (see Figure \ref{fig:fair_abstract_box_plot}).
Intuitively, one can argue that having a single threshold $\tau$ that holds Premises \ref{pre:A} and \ref{pre:B} can be problematic.

\begin{figure}[ht]
     \begin{center}

        \includegraphics [width=26em] {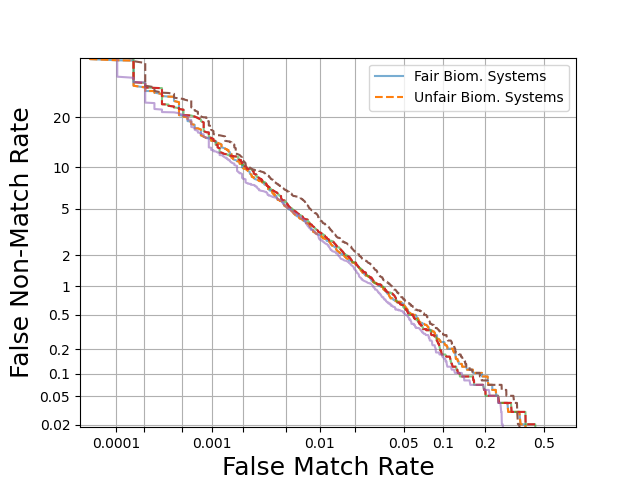}
        \caption{DET curves for the canonical fair and unfair synthetic verification systems. It can be observed that analysing this curves gives a false impression that the unfair synthetic verification system is fair.}
        \label{fig:roc_fair_unfair}
     \end{center}
     \vspace{-1em}
\end{figure}

Let us now test FDR using these two theoretical systems\footnote {This example is available in the following link: https://github.com/tiagofrepereira2012/fdr/}.
Table \ref{tb:fair} presents $\text{FNMR}(\tau)$, $\text{FMR}(\tau)$ and $\text{FDR}(\tau)$ for different values of $\tau$ of the fair synthetic biometric system presented in Figure \ref{fig:fair_abstract_box_plot}.
In this experiment $\tau =\text{FMR}_{x}(\tau)$ where $x$ varies from $10^{-3}$ to $10^{-6}$.
It is possible to observe that $\text{FDR}(\tau)$ is stable and higher than $0.99$ for all values of $x$, which indicates non-discrepant behavior concerning these abstract demographics.
To analyse the other side of the spectrum, Table \ref{tb:unfair} presents $\text{FNMR}(\tau)$, $\text{FMR}(\tau)$ and $\text{FDR}(\tau)$ for different values of $\tau$ of the unfair synthetic biometric system presented in Figure \ref{fig:unfair_abstract_box_plot}.
The values of $\tau$ are set in the same way as in the previous experiment.
It is possible to observe that $\text{FDR}(\tau)$ is consistently higher for the fair biometric system than with the unfair one, which indicates consistency in this figure of merit in the evaluation of demographic differentials.
On the other hand, Figure \ref{fig:roc_fair_unfair} shows the DET curves of these two synthetic examples and the three demographics.
It is possible to observe that all six DET curves present the same trends, overlapping each other.
This example gives the false impression that the unfair synthetic verification system is fair, which is not as we could spot with the FDR.

\begin{figure}[ht]
     \begin{center}

        \includegraphics [width=26em] {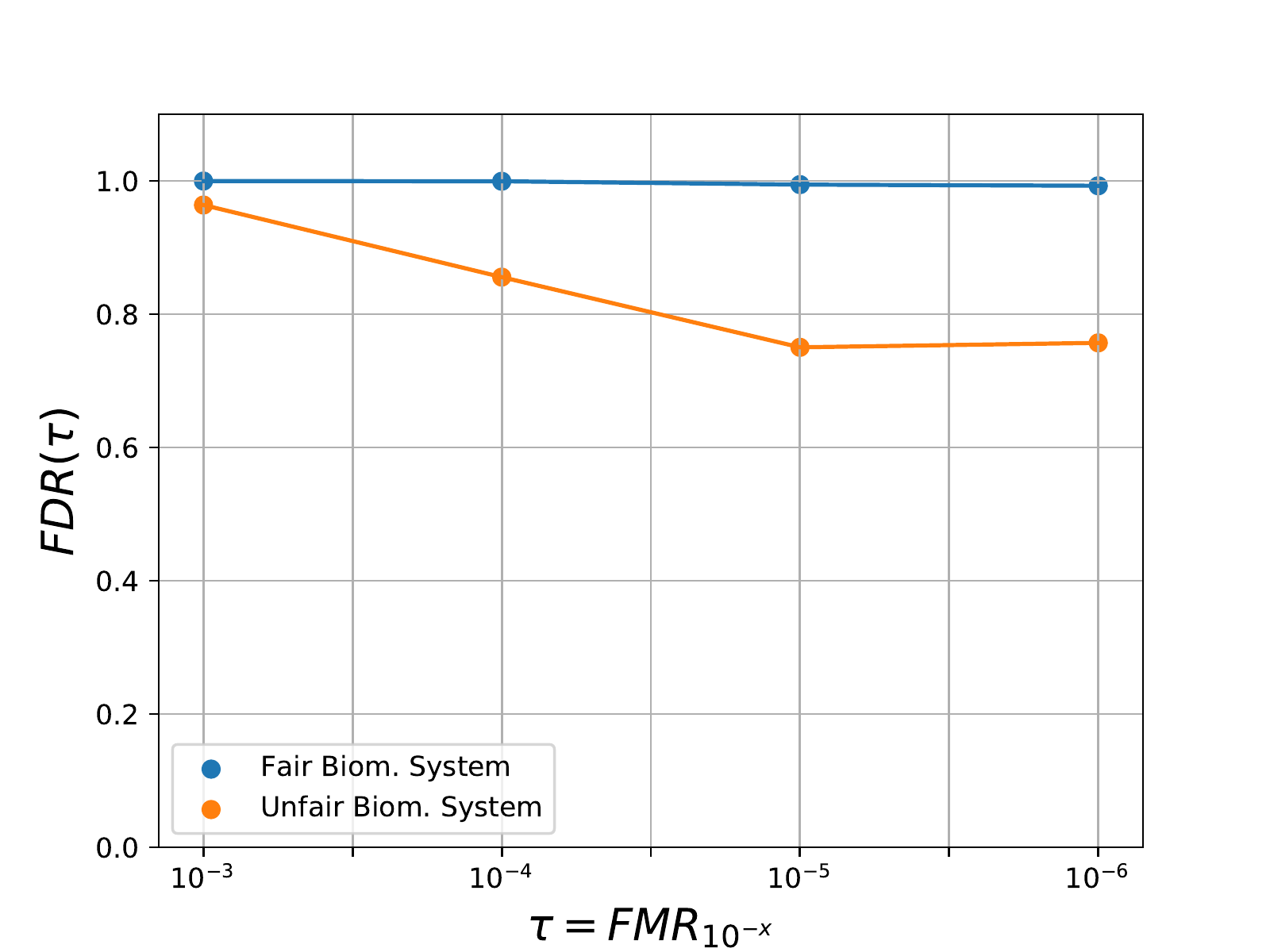}
        \caption{FDR as a function of $x$ from two synthetic biometric systems from Figures \ref{fig:fair_abstract_box_plot} and \ref{fig:unfair_abstract_box_plot}.}
        \label{fig:fair_metric}
     \end{center}
     \vspace{-1em}
\end{figure}

It is possible to plot FDR as a function of $x$ (or $\tau$) to compare two biometric systems.
Figure \ref{fig:fair_metric} presents how the two biometric systems can be compared under this figure of merit.
It is possible to observe that FDR is stable for all values of $x$ for the fair biometric system.
FDR substantially decreases once $x$ decreases (when fewer false-matches are allowed) for the unfair biometric system.
Another way to establish a comparison between two systems concerning demographic discrepancies is by analyzing the Area Under FDR.
For a given range of $\tau$ (estimated by using $x$), the Area Under FDR can be calculated by merely integrating the $\text{FDR}(\tau)$ over $x$.
The value of $x$ is scaled from 0 to 1, so Area Under FDR is bounded to this range.
However, by scaling it, the range of $x$ has to be reported.
Hence, only Area Under FDR whose range of $x$ matches can be fairly compared.
Using this figure of merit, it is also possible to observe that the system that was intuitively considered as fair (see Figure \ref{fig:fair_abstract_box_plot}) presents a Area Under FDR of $0.99$ and the one that was intuitively considered as unfair presents a Area Under FDR of $0.88$ (see Figure \ref{fig:unfair_abstract_box_plot}).


\subsection{The role of alpha}

The hyper-parameter $\alpha \in [0,1]$ in equation \ref{eq:fairness} has a crucial role in the computation of $\text{FDR}(\tau)$.
As previously mentioned, it controls the weight of False Matches in the FDR computation.
Such value is a business/application decision.
For instance, a bank that deploys a biometric verification system in an ATM might prefer to favor parity towards False Non-Matches and, for this reason, $\alpha$ can assume low values.
On the other hand, in a border control scenario, where false matches are critical, decision-makers might favor False-Matches' parity.
Hence, $\alpha$ should be high.

\begin{figure}[ht]
     \begin{center}

        \includegraphics [width=26em] {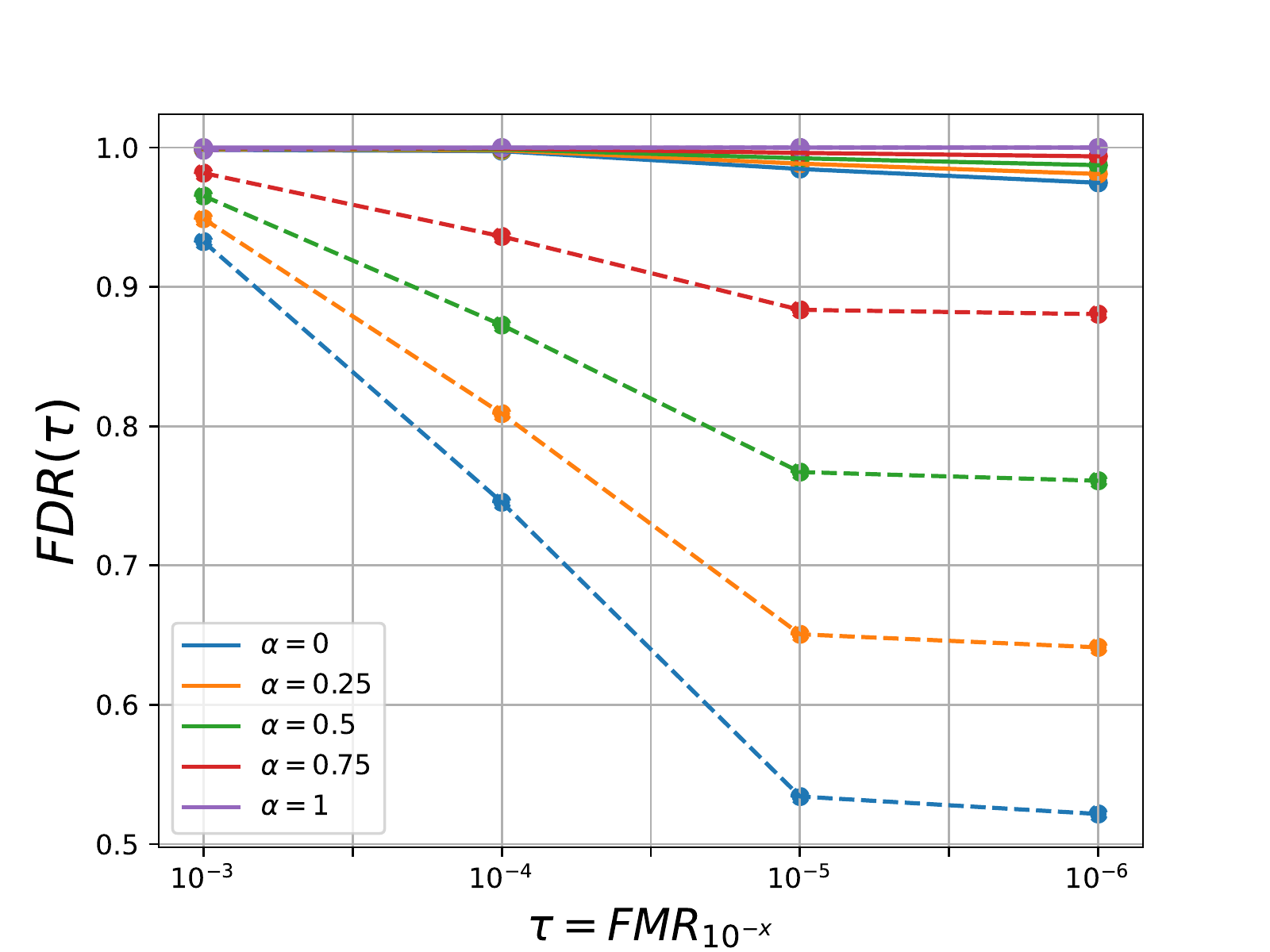}
        \caption{$\text{FDR}(\tau)|\alpha$ for different values of $\alpha$ (see equation \ref{eq:fairness}). Solid lines represent the $\text{FDR}(\tau)|\alpha$ for the synthetic fair verification system. Dashed lines represent the $\text{FDR}(\tau)|\alpha$ for the synthetic unfair verification system.}
        \label{fig:fpr_over_alpha}
     \end{center}
     \vspace{-1em}
\end{figure}

Figure \ref{fig:fpr_over_alpha} shows the $\alpha$ trade-off between the two synthetic systems; the fair ones are represented by the solid lines and the unfair by the dashed lines.
It is possible to observe that the fair system presents a $\text{FDR}(\tau) \sim 0.98$ no matter the value of $\alpha$.
For the unfair system, $\text{FDR}(\tau)$ presents a stepper decay once $\alpha$ decreases.
This also can be seen via the Area Under FDR.
As can be noticed in Table \ref{tb:aufdr_over_alpha}, for the unfair biometric system, the Area Under FDR substantially changes once $\alpha$ changes.

\begin{table}[ht]
   \begin{center}
   
   \caption{Area Under the FDR for different values of $\alpha$ for $x$ varying from $10^{-3}$ to $10^{-6}$.}

    \begin{tabular}{r|rr|}

      \cline{2-3}
       &  Fair &   Unfair  \\ \hline
	
       \multicolumn{1}{|l|}{$\alpha = 0$}    & 0.99 & 0.81 \\
       \multicolumn{1}{|l|}{$\alpha = 0.25$} & 0.99 & 0.86 \\
       \multicolumn{1}{|l|}{$\alpha = 0.5$}  & 0.99 & 0.90 \\
       \multicolumn{1}{|l|}{$\alpha = 0.75$} & 0.99 & 0.95 \\
       \multicolumn{1}{|l|}{$\alpha = 1$}    & 0.99 & 0.98 \\
      \hline
      
   \end{tabular}

  \label{tb:aufdr_over_alpha}
  \end{center}  
\end{table}



\subsection{The role of epsilon}

Another hyper-parameter in this figure of merit is the values that $\epsilon$ can assume.
As mentioned in equation \ref{eq:equality}, this value is supposed to define what is fair and what is not with respect to $\text{FDR}(\tau)$.
To the best of our knowledge, there is no reference value that we can rely upon.
In some particular cases, there are some guidelines.
For instance, as mentioned in \cite[(sec 2, p.19)]{barocas-hardt-narayanan}, the U.S Equal Employment Opportunity Commission\footnote{https://www.eeoc.gov/laws/guidance/employment-tests-and-selection-procedures} states that a disparate behavior between two groups occurs if the probability of selection between them differ from more than 20\%.
We can reasonably agree that having \textbf{$\epsilon \sim 0.20$} in equation \ref{eq:equality} is not realistic for a biometric verification system.

This work will not draw a line to define what is fair and what is not for biometric verification systems.
As mentioned before, there is no legal or technical basis for such, and the ones that do exists are not suitable for biometrics.
Instead, we will use both FDR and Area Under FDR to compare different biometric verification systems in terms of demographic differentials.



\section{Face Verification use case}
\label{sec:fv_usecase}

In this section, a case study of the Fairness Discrepancy Rate is presented using three FR systems based on DCNNs and one Commercial-Off-The-Shelf System (COTS).
The \textbf{first} DCNN system is based on the Inception-Resnet v2 architecture \cite{szegedy2016inception}.
This model was trained with the MS-Celeb-1M dataset using a joint loss function combining the cross-entropy loss and center loss.
Check in \cite[p.147]{deFreitasPereira2019learning} more details on how this DCNN was trained.
The \textbf{second} DCNN is the ArcFace model \cite{deng2019arcface} from InsightFace (ArcFace-InsightFace)\footnote{https://github.com/deepinsight/insightface}.
This model is based on Resnet-100 architecture \cite{he2016deep}, and trained using the ArcFace loss.
The \textbf{third} is Resnet-50 based architecture \cite{he2016deep} trained using the ArcFace loss.
Such a model was trained using the VGG2Face dataset\cite{cao2018vggface2}.
More details on this DCNN was trained can be found in \footnote{https://gitlab.idiap.ch/bob/bob.bio.face}.
For these three biometric systems, comparisons between samples are made with the embeddings of each DCNN using the cosine distance metric.
Given the embeddings $e$ and $p$ for enrollment and probing, respectively, the distance $s(e,p)$ is given by:
\begin{equation}
s(e ,p) = 1 - \frac{e \cdot p}{\lVert e \rVert \cdot \lVert p\rVert } 
\label{eq:cos_sim}
\end{equation}

Finally, the \textbf{fourth} face verification system a COTS developed by RankONE \footnote{https://www.rankone.io version 1.22.1 }.
\subsection{Dataset setup}

Several datasets are publicly available in the literature with privacy-sensible attributes where FR systems assessments can be carried out.
The most recent ones available are based on images from celebrities scraped from the web, such as Racial Faces in the Wild (RFW) \cite{wang2019racial}, Balanced Faces in the Wild \cite{robinson2020face} and, IARPA Janus Benchmark C (IJB-C) \cite{maze2018iarpa}.
Although all the aforementioned datasets contain meta-information that can be used to assess demographic differentials, they were captured in the so-called ``in a wild'' conditions.
Several factors of variation, such as pose, illumination, occlusion, and image quality, can play a role in the final recognition rates, which might interfere with this initial FDR assessment.
Since this is the first work with this figure of merit, we have focused on three datasets where capture conditions are ``less in the wild'' and whose demographic attributes are available.







NIST developed the MEDS II database to support and assists their biometrics evaluation program.
Five hundred eighteen identities compose it from both men/women (labeled as M and F) and five different race annotations, and they are Asian, Black, American Indian, Unknown and White (labeled as A, B, I, U, and W).
Unfortunately, the distribution of gender and race is hugely unbalanced.
Furthermore, only 256 subjects have more than one image sample (obviously, it is impossible to do a biometric evaluation with one sample per subject).
For this reason, we have performed our evaluation in a subset of this dataset, which is composed only of 223 subjects composed of White and Black men only (where we have 109 white subjects and 114 back subjects).
More details about the organization of this evaluation setup can be found in its website$^13$.

Although dating from 2008, the MORPH dataset is getting some traction recently (\cite{vangara2019characterizing,cavazos2019accuracy,krishnapriya2020issues}) mostly because of its richness concerning sensitive attributes.
It is composed of 55,000 samples from 13,000 subjects from men and women and five race clusters (called ancestry), and they are the following: African, European, Asian, Hispanic, and Others.
More details about the organization of this evaluation setup can be found  in its website$^{13}$.


\begin{table}[ht]
   \begin{center}

   \caption{MEDS II - ArcFace-InsightFace: $\text{FNMR}(\tau)$, $\text{FMR}(\tau)$, and $\text{FDR}(\tau)$ per demographic (Demog.) in the test set. These figures of merit are fragmented by the race of the samples used for enrollment and the race of the samples used for probe (``(e-p)'' in the table.)\textsuperscript{*}.}

    \begin{tabular}{|r||rrrrr|}
      \hline
  
       $\tau=FMR_{10^{x}}$  &  $10^{-1}$ &   $10^{-2}$ &   $10^{-3}$ &   $10^{-4}$ &   $10^{-5}$  \\
	
      \hline
      Demog (e-p) & \multicolumn{5}{c|}{$\text{FMR}_x(\tau)$} \\
      \hline
  White - White & \colorMyCellMEDS{0.106}  &  \colorMyCellMEDS{0.008} &  \colorMyCellMEDS{0.001}  &     \colorMyCellMEDS{0}    &    \colorMyCellMEDS{0} \\
  White - Black & \colorMyCellMEDS{0.058}  &  \colorMyCellMEDS{0.002} &  \colorMyCellMEDS{0     } &     \colorMyCellMEDS{0}    &    \colorMyCellMEDS{0} \\
  Black - White & \colorMyCellMEDS{0.068}  &  \colorMyCellMEDS{0.001} &  \colorMyCellMEDS{0     } &     \colorMyCellMEDS{0}    &    \colorMyCellMEDS{0} \\
  Black - Black & \colorMyCellMEDS{0.162}  &  \colorMyCellMEDS{0.026} & \colorMyCellMEDS{0.003}   &     \colorMyCellMEDS{0}    &    \colorMyCellMEDS{0} \\
      \hline
      & \multicolumn{5}{c|}{$\text{FNMR}_x(\tau)$} \\
      
      \hline
  White - White & \colorMyCellMEDS{0} &   \colorMyCellMEDS{0} &    \colorMyCellMEDS{0}  &   \colorMyCellMEDS{0}  & \colorMyCell{0.016} \\
  Black - Black & \colorMyCellMEDS{0} &   \colorMyCellMEDS{0} &    \colorMyCellMEDS{0}  &   \colorMyCellMEDS{0}  & \colorMyCell{0} \\

       \hline
        $FDR(\tau)$ & 0.972  &  0.991  &   0.995  &  0.998 &  0.995 \\
       \hline        
      
   \end{tabular}

  \label{tb:meds}
  \end{center}  
  
  \small\textsuperscript{\textsuperscript{*}In this example $\tau=\text{FMR}^{\text{dev}}_{x}$ where \textbf{dev} is the development-set.}
  \small\textsuperscript{$\text{FMR}(\tau)$, $\text{FNMR}(\tau)$ and $FDR(\tau)$ are reported using the test-set}
  
\end{table}

The MOBIO dataset is a video database containing bi-modal data (face/speaker).
One hundred fifty-two people compose it (split in the two genders \textbf{male} and \textbf{female}), mostly Europeans, split into five sessions (few weeks time lapse between sessions).
The database was recorded using two types of mobile devices: mobile phones (NOKIA N93i) and laptop computers(standard 2008 MacBook).   In this paper, we only use mobile phone data.
As with other datasets, its evaluation protocol is also published as a python package\footnote{https://gitlab.idiap.ch/bob/bob.db.mobio}.


\subsection{Experiments}

In this section, we discuss the demographic differentials of the four FR systems using the Fairness Discrepancy Rate.
Each one of the following subsections discusses each database in isolation.
In each of the experiments, both False Matches and False Non-Matches have the same weight; therefore, $\alpha$ is equal to $0.5$.
As aforementioned, we will not set a value for $\epsilon$; instead, we will use FDR and Area Under FDR to compare different biometric verification systems concerning demographic discrepancies.

\subsection{MEDS II database (Fairness with respect to race)}

\begin{figure}[ht]
     \begin{center}

        \includegraphics [width=26em,page=1] {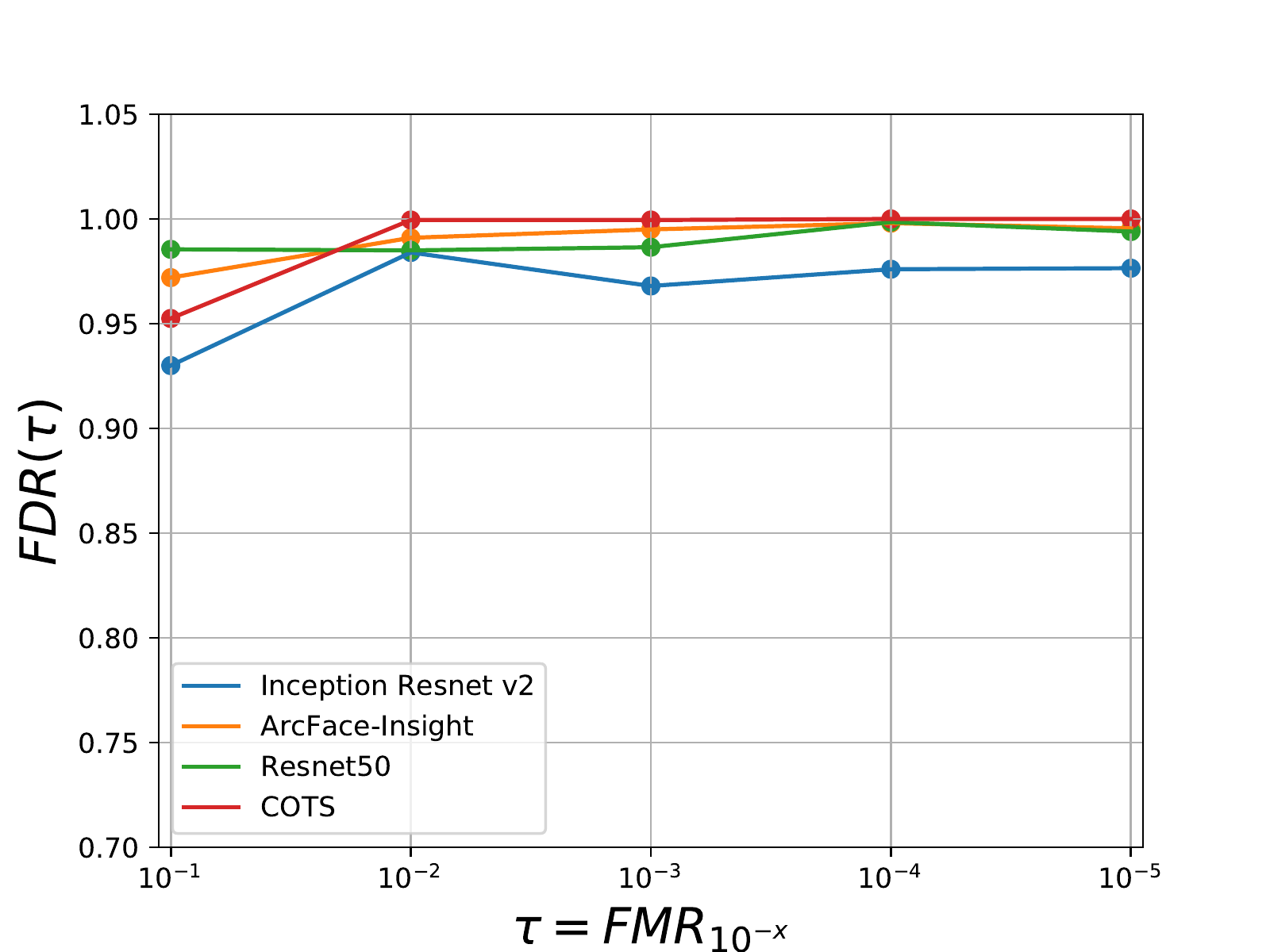}
        \caption{MEDS II: Fairness Discrepancy Rate of different face verification systems for different decision thresholds}
        \label{fig:meds_fdr}
     \end{center}
     \vspace{-1em}
\end{figure}

Table \ref{tb:meds} presents the $\text{FMR}(\tau)$, $\text{FNMR}(\tau)$ and $\text{FDR}(\tau)$ in the test set for the ArcFace-InsightFace system.
For the sake of brevity, we present only this system in this comprehensive manner.
Please, check the supplementary material to have information about the other systems.
In this experiment, $\tau$ was set at different operational points in the impostor score distribution from an independent set (development set in this case).
It is worth noting that such impostor score distribution contains samples from all races, which is the closest scenario to operational conditions, where one single threshold has to be fair to all demographic groups.
Both $\text{FMR}_x(\tau)$ and $\text{FNMR}_x(\tau)$ tables are fragmented by demographics (race in this case).
Hence, in Table \ref{tb:meds}, ``White - White'' means biometric references from White subjects compared with probe samples from White subjects, and so on.

In terms of $\text{FMR}_x(\tau)$ it is possible to observe that for $x=10^{-1}$ and $x=10^{-2}$ ($\tau=\text{FMR}_{10^{-1}}$ or $\tau=\text{FMR}_{10^{-2}}$) the face verification system tends to have more false alarms for comparison between biometric references and probes from Black subjects.
We can also observe the number of false alarms using the comparison scores from different races (e.g. ``White-Black'' and ``Black-White'') is substantially lower than the number of false alarms using the pair gallery-probe from the same race.
This behavior was observed and reported in several publications \cite{grother2019face,howard2019effect,michalski2018impact}. 
In terms of $\text{FNMR}_x(\tau)$, it is possible to notice that such a system tends to reject more White subjects than Black for $x > 10^{-4}$.

Figure \ref{fig:meds_fdr} presents the FDR plot of the four different biometric systems covering the same decision thresholds showed in Table \ref{tb:meds}.
We can observe that all the four FR systems tend to have more discrepancies between $x=10^{-1}$ and $x=10^{-2}$.
This reflects the amount of the false-alarms raised in these operational points (see Table \ref{tb:meds}).
The system based on Inception-Resnet-v2 presents more discrepancies than the other ones for all operational points.
This possibly indicates that the recent state-of-the-art approaches are naturally fairer than the past approaches, even if fairness constraints are not considered in the current approaches (see \cite{deng2019arcface}).

Table \ref{tb:meds_aufdr} (a) presents the Area Under FDR  (varying  from $10^{-1}$ to $10^{-5}$) of every  biometric verification system.
With this summarized figure of merit, we can spot some of the trends spotted with Figure \ref{tb:meds}.
Furthermore, we can observe that the current state-of-the-art FR approaches are fairer than the evaluated COTS.

\begin{table}[ht]
   \begin{center}
   
   \caption{Area Under the Fairness Discrepancy Rate for (a) MEDS II, (b) MORPH and (c) MOBIO}

    \begin{tabular}{r|c|c|c|}

      \cline{2-4}
       &  \multicolumn{3}{c|}{FDR AUC}   \\ \cline{2-4}
       &  \multicolumn{1}{c|}{(a) MEDS II}                                    & (b) MORPH      & (c) MOBIO   \\ \hline
       \multicolumn{1}{|c|}{Inception-Resnet-v2}         & 0.959               & 0.82           & 0.93 \\
       \multicolumn{1}{|c|}{ArcFace-InsightFace}         & \textbf{0.991}      & \textbf{0.938} & 0.96  \\
       \multicolumn{1}{|c|}{Resnet50}                    & 0.985               &  0.926         & 0.92 \\
       \multicolumn{1}{|c|}{COTS}                        & 0.97                &  0.81          &  \textbf{0.97} \\
      \hline
      
   \end{tabular}

  \label{tb:meds_aufdr}
  \end{center}  
\end{table}

\begin{figure}[ht]
     \begin{center}

        \includegraphics [width=26em,page=2] {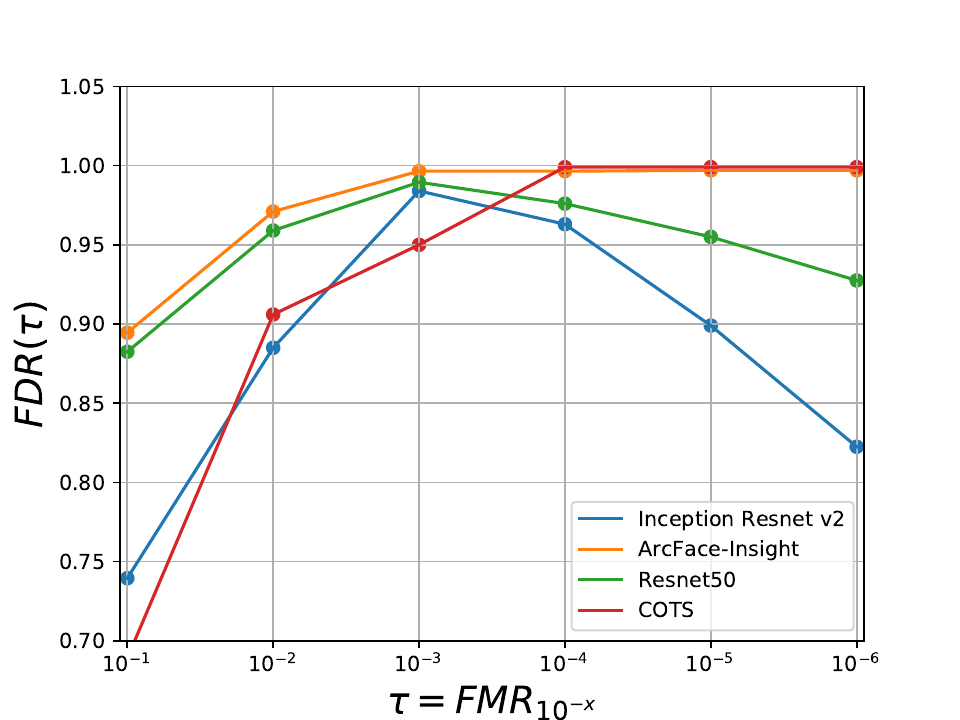}
        \caption{Morph: Fairness Discrepancy Rate of different face verification systems for different decision thresholds}
        \label{fig:morph_fdr}
     \end{center}
     \vspace{-1em}
\end{figure}

\begin{table*}[ht]
   \begin{center}

   \caption{MORPH - ArcFace-InsightFace: $\text{FNMR}(\tau)$, $\text{FMR}(\tau)$, and $\text{FDR}(\tau)$ per demographic (Demog.) in the test set. These figures of merit are fragmented by the race of the samples used for enrollment and the race of the samples used for probe (``(e-p)'' in the table.)\textsuperscript{*}}


    \begin{tabular}{|r||cccccc|}
      \hline
  
       $\tau=FMR_{10^{x}}$  &  $10^{-1}$ &   $10^{-2}$ &   $10^{-3}$ &   $10^{-4}$ &   $10^{-5}$ &   $10^{-6}$ \\
	
      \hline
      Demog. (e-p) & \multicolumn{6}{c|}{$\text{FMR}_x(\tau)$} \\
      \hline
Asian - Asian & \colorMyCellMORPH{0.314}  &  \colorMyCellMORPH{0.066}  &  \colorMyCellMORPH{0}     &   \colorMyCellMORPH{0}     & \colorMyCellMORPH{0}   &  \colorMyCellMORPH{0}  \\
Asian - Black & \colorMyCellMORPH{0.076}  &  \colorMyCellMORPH{0.003}  &  \colorMyCellMORPH{0}     &    \colorMyCellMORPH{0}    & \colorMyCellMORPH{0}   &  \colorMyCellMORPH{0}  \\
Asian - Hisp. & \colorMyCellMORPH{0.108}  &  \colorMyCellMORPH{0.009}  &  \colorMyCellMORPH{0}     &   \colorMyCellMORPH{0}     & \colorMyCellMORPH{0}   &  \colorMyCellMORPH{0}  \\
Asian - White & \colorMyCellMORPH{0.063}  &  \colorMyCellMORPH{0.002}  &  \colorMyCellMORPH{0}     &   \colorMyCellMORPH{0}     & \colorMyCellMORPH{0}   &  \colorMyCellMORPH{0}  \\
Black - Asian & \colorMyCellMORPH{0.068}  &  \colorMyCellMORPH{0.002}  &  \colorMyCellMORPH{0}     &   \colorMyCellMORPH{0}     & \colorMyCellMORPH{0}   &  \colorMyCellMORPH{0}  \\
Black - Black & \colorMyCellMORPH{0.148}  &  \colorMyCellMORPH{0.019}  &  \colorMyCellMORPH{0}     &   \colorMyCellMORPH{0}     & \colorMyCellMORPH{0}   &  \colorMyCellMORPH{0}  \\
Black - Hisp. & \colorMyCellMORPH{0.067}  &  \colorMyCellMORPH{0}      &  \colorMyCellMORPH{0}     &   \colorMyCellMORPH{0}     & \colorMyCellMORPH{0}   &  \colorMyCellMORPH{0}  \\
Black - White & \colorMyCellMORPH{0.053}  &  \colorMyCellMORPH{0}      &  \colorMyCellMORPH{0}     &   \colorMyCellMORPH{0}     & \colorMyCellMORPH{0}   &  \colorMyCellMORPH{0}  \\
Hisp. - Asian & \colorMyCellMORPH{0.102}  &  \colorMyCellMORPH{0.006}  &  \colorMyCellMORPH{0}     &   \colorMyCellMORPH{0}     & \colorMyCellMORPH{0}   &  \colorMyCellMORPH{0}  \\
Hisp. - Black & \colorMyCellMORPH{0.066}  &  \colorMyCellMORPH{0}      &  \colorMyCellMORPH{0}     &   \colorMyCellMORPH{0}     & \colorMyCellMORPH{0}   &  \colorMyCellMORPH{0}  \\
Hisp. - Hisp. & \colorMyCellMORPH{0.217}  &  \colorMyCellMORPH{0.005}  &  \colorMyCellMORPH{0.001} &   \colorMyCellMORPH{0}     & \colorMyCellMORPH{0}   &  \colorMyCellMORPH{0}  \\
Hisp. - White & \colorMyCellMORPH{0.063}  & \colorMyCellMORPH{0.001}   &  \colorMyCellMORPH{0}     &   \colorMyCellMORPH{0}     & \colorMyCellMORPH{0}   &  \colorMyCellMORPH{0}  \\
White - Asian & \colorMyCellMORPH{0.064}  &  \colorMyCellMORPH{0}      &  \colorMyCellMORPH{0}     &   \colorMyCellMORPH{0}     & \colorMyCellMORPH{0}   &  \colorMyCellMORPH{0}  \\
White - Black & \colorMyCellMORPH{0.054}  &  \colorMyCellMORPH{0}      &  \colorMyCellMORPH{0}     &   \colorMyCellMORPH{0}     & \colorMyCellMORPH{0}   &  \colorMyCellMORPH{0}  \\
White - Hisp. & \colorMyCellMORPH{0.063}  &  \colorMyCellMORPH{0.}     &  \colorMyCellMORPH{0}     &   \colorMyCellMORPH{0}     & \colorMyCellMORPH{0}   &  \colorMyCellMORPH{0}  \\
White - White & \colorMyCellMORPH{0.103}  &  \colorMyCellMORPH{0.008}  &  \colorMyCellMORPH{0}     &   \colorMyCellMORPH{0}     & \colorMyCellMORPH{0}   &  \colorMyCellMORPH{0}  \\

      \hline
      & \multicolumn{6}{c|}{$\text{FNMR}_x(\tau)$} \\
      
      \hline

Asian - Asian & \colorMyCellMORPH{0}  & \colorMyCellMORPH{0}      & \colorMyCellMORPH{0}   &  \colorMyCellMORPH{0}      & \colorMyCellMORPH{0}      &   \colorMyCellMORPH{0}      \\
Black - Black & \colorMyCellMORPH{0}  & \colorMyCellMORPH{0}      & \colorMyCellMORPH{0}   &  \colorMyCellMORPH{0}      & \colorMyCellMORPH{0}      &   \colorMyCellMORPH{0.001}  \\
Hisp. - Hisp. & \colorMyCellMORPH{0}  & \colorMyCellMORPH{0}      & \colorMyCellMORPH{0}   &  \colorMyCellMORPH{0.006}  & \colorMyCellMORPH{0.006}  &   \colorMyCellMORPH{0.006}  \\
White - White & \colorMyCellMORPH{0}  & \colorMyCellMORPH{0}      & \colorMyCellMORPH{0}   &  \colorMyCellMORPH{0}      & \colorMyCellMORPH{0.002}  &   \colorMyCellMORPH{0.005}  \\

       \hline
        $FDR(\tau)$ & 0.8945  & 0.9965  &  0.9965  &   0.997  &  0.899 &  0.997 \\
       \hline        
      
   \end{tabular}


  \label{tb:morph}
  \end{center}  
  
  \small\textsuperscript{\textsuperscript{*}In this example $\tau=\text{FMR}^{\text{dev}}_{x}$ where \textbf{dev} is the development-set.}
  \small\textsuperscript{$\text{FMR}(\tau)$, $\text{FNMR}(\tau)$ and $FDR(\tau)$ are reported using the test-set}
  
\end{table*}

\subsection{MORPH database (Fairness with respect to race)}

Table \ref{tb:morph} presents the $\text{FMR}(\tau)$, $\text{FNMR}(\tau)$ and $\text{FDR}(\tau)$ in the test set (Male subjects only) for the ArcFace-InsightFace verification system.
As with the last section, for brevity, only this system will be presented in this extensive manner.
In this experiment $\tau$ was set at different operational points in the impostor score distribution from the development set.
Both $\text{FMR}_x(\tau)$ and $\text{FNMR}_x(\tau)$ tables are fragmented by demographics (race in this case) in the same manner as in the previous experiment.
However, in this one, we have four demographic groups, and they are the following: Asian, Black, Hispanic, and White (samples labeled as ``Others'' were left aside).

In terms of $\text{FMR}_x(\tau)$ it is possible to observe that from $x=10^{-1}$ to $x=10^{-2}$ (from $\tau=\text{FMR}_{10^{-1}}$ to $\tau=\text{FMR}_{10^{-2}}$) the face verification system tends to have more false alarms for comparisons between biometric references and probes from Hispanic, Asian and Black subjects.
Interesting to observe that for $x=10^{-1}$, a significant amount of false alarms are observed between Asian biometric references with Hispanic Probes and vice-versa.
In terms of $\text{FNMR}_x(\tau)$, it is possible to notice that such a system tends to reject more Hispanic subjects from $x \geq 10^{-4}$ and more White subjects from $x \geq 10^{-5}$.

Figure \ref{fig:morph_fdr} presents the FDR plot of the four different biometric verification systems covering the same decision thresholds showed in Table \ref{tb:morph}.
We can observe that the four face verification systems present demographic differentials from $x=10^{-1}$ to $x=10^{-3}$ mostly due to false matches.
From $x=10^{-3}$ to $x=10^{-6}$ the ArcFace-InsightFace and the COTS presents very little demographic differentials.
Both operate at $FDR(\tau) \geq 0.99$.
Demographic differentials from Resnet50 and Inception-Resnet-v2 substantially increase from $x > 10^{-3}$, mostly due to the differences in false non-matches.

Table \ref{tb:meds_aufdr} (b) presents a full picture of the above observations with the Area Under FDR ($x$ varying from $10^{-1}$ to $10^{-6}$)\footnote{In this experiment, we have enough scores to place a $\tau$ at $10^{-6}$} of every biometric verification system.
We can observe, in this experiment, that the ArcFace-InsightFace is the fairest FR system, followed by the Resnet50, both based on the ArcFace loss.
COTS and the Inception-Resnet-v2 systems present similar $FDR$ ($\approx 0.82$).

   


	
      


\subsection{MOBIO Database (Fairness with respect to gender)}

Table \ref{tb:mobio} presents the $\text{FMR}(\tau)$, $\text{FNMR}(\tau)$ and $\text{FDR}(\tau)$ in the test set for the ArcFace-InsightFace system.
MOBIO dataset is composed basically of Caucasians, and for that reason, this experiment focuses on gender differentials only.
Hence, $\text{FMR}_x(\tau)$ and $\text{FNMR}_x(\tau)$ tables are fragmented by gender in the same manner as in the previous experiments.
In this setup, $\tau$ is set at different operational points in an independent zeroth-effort impostor score distribution (from the development set).

\begin{table}[ht]
   \begin{center}

   \caption{MOBIO - ArcFace-InsightFace: $\text{FNMR}(\tau)$, $\text{FMR}(\tau)$, and $\text{FDR}(\tau)$ per gender in the test set. These figures of merit are fragmented by the gender of the samples used for enrollment and the race of the samples used for probe (``(e-p)'' in the table.)\textsuperscript{*}.}

    \begin{tabular}{|r||rrrrr|}
      \hline
  
      $\tau=FMR_{10^{x}}$  &  $10^{-1}$ &   $10^{-2}$ &   $10^{-3}$ &   $10^{-4}$ &   $10^{-5}$  \\
	
      \hline
      Demog.(e-p) & \multicolumn{5}{c|}{$\text{FMR}_x(\tau)$} \\
      \hline
  Male-Male     & \colorMyCellMOBIO{0.077}  & \colorMyCellMOBIO{0.006}     &   \colorMyCellMOBIO{0}      & \colorMyCellMOBIO{0}     & \colorMyCellMOBIO{0}\\
  Male-Female   & \colorMyCellMOBIO{0.052}  & \colorMyCellMOBIO{0.001}     &   \colorMyCellMOBIO{0}      & \colorMyCellMOBIO{0}     & \colorMyCellMOBIO{0}  \\
  Female-Male   & \colorMyCellMOBIO{0.043}  & \colorMyCellMOBIO{0.001}     &   \colorMyCellMOBIO{0}      &  \colorMyCellMOBIO{0}    & \colorMyCellMOBIO{0} \\
  Female-Female & \colorMyCellMOBIO{0.235}  & \colorMyCellMOBIO{0.027}     &   \colorMyCellMOBIO{0.004}  &  \colorMyCellMOBIO{0}    & \colorMyCellMOBIO{0}\\
      \hline
      & \multicolumn{5}{c|}{$\text{FNMR}_x(\tau)$} \\
      
      \hline
  Male-Male     & \colorMyCellMOBIO{0.001} & \colorMyCellMOBIO{0.001} &  \colorMyCellMOBIO{0.001} & \colorMyCellMOBIO{0.001}  &  \colorMyCellMOBIO{0.002} \\
  Female-Female & \colorMyCellMOBIO{0}     &  \colorMyCellMOBIO{0}    & \colorMyCellMOBIO{0}      & \colorMyCellMOBIO{0}      &  \colorMyCellMOBIO{0} \\

       \hline
        $FDR(\tau)$ & 0.9205 &  0.989  &  0.997  &   0.999  &  0.999 \\
       \hline        
      
   \end{tabular}

  \label{tb:mobio}
  \end{center}
    \small\textsuperscript{\textsuperscript{*}In this example $\tau=\text{FMR}^{\text{dev}_{x}}$ where \textbf{dev} is the development-set.}
  \small\textsuperscript{$\text{FMR}(\tau)$, $\text{FNMR}(\tau)$ and $FDR(\tau)$ are reported using the test-set}

\end{table}

In terms of $\text{FMR}_x(\tau)$ it is possible to notice that from $x=10^{-1}$ to $x=10^{-2}$ (from $\text{FMR}_{10^{-1}}(\tau)$ or $\text{FMR}_{10^{-2}}(\tau)$) the face verification system tends to have more false alarms for comparison between biometric references and probes from female subjects.
We can also observe the number of false alarms using the comparison scores from different genders (e.g. ``Male-Female'' and ``Female-Male'') is substantially lower than the number of false alarms from comparison scores from the same gender.
These are the same trends observed before with race and, as already mentioned, the same trends observed in the literature.
In terms of $\text{FNMR}_x(\tau)$ such a system tends to slightly reject more Male subjects at $x=10^{-5}$.

\begin{figure}[ht]
     \begin{center}

        \includegraphics [width=26em,page=3] {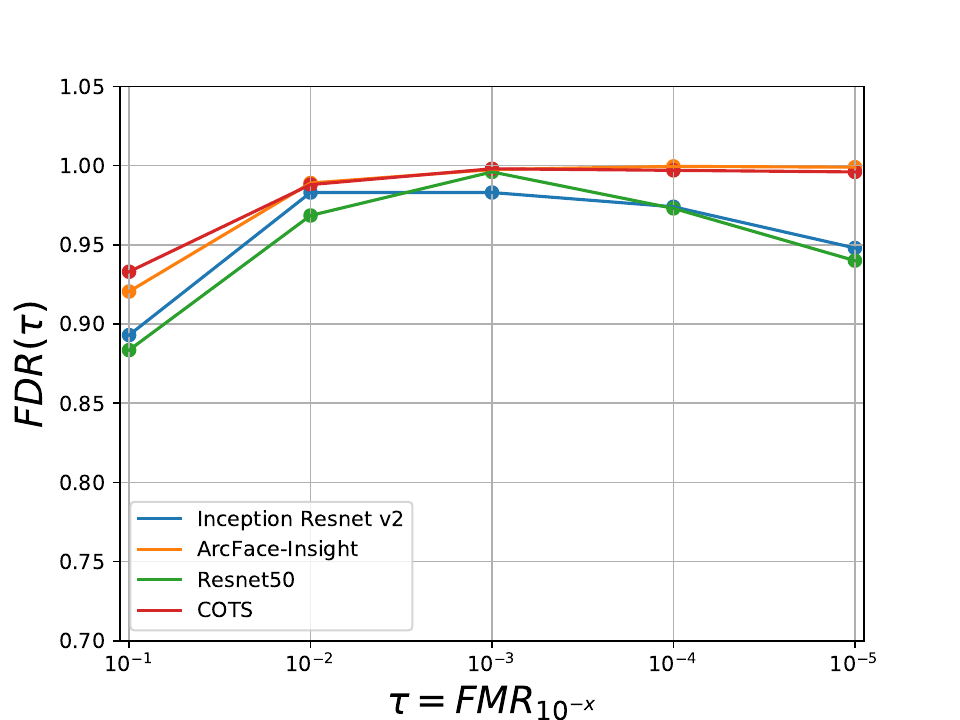}
        \caption{MOBIO: Fairness Discrepancy Rate of different face verification systems for different decision thresholds}
        \label{fig:mobio_fdr}
     \end{center}
     \vspace{-1em}
\end{figure}

Figure \ref{fig:mobio_fdr} presents the overall picture about demographic differentials using FDR.
We can observe that the four face verification systems present demographic differentials from $x=10^{-1}$ to $x=10^{-2}$ mostly due to false matches.
As in the previous experiment, the ArcFace-InsightFace and the COTS  presents very little demographic differentials from $x=10^{-3}$ to $x=10^{-5}$; with both operating at $FDR(\tau) \geq 0.99$.
The FDR for the systems based on Inception-Resnet-v2 and Resnet50 decreases from $x \geq 10^{-3}$, mostly due to false non-matches.
Finally, Table \ref{tb:meds_aufdr} (c) presents the Area Under FDR.
We can observe that, in this experiment, the COTS is the fairest FR system, followed by the ArcFace-InsightFace by a short margin.


\vspace{-1.5em}

\subsection{Discussion}

This section presented a case study using the proposed Fairness Discrepancy Rate to assess demographic differentials.
Experiments with three open-source FR baselines based on DCNNs and one COTS system were used along with three databases where gender and racial differentials were studied.
We could notice that  FDR could summarize and compare the demographic differentials concerning FMR and FNMR between several FR systems.
Furthermore, the Area Under FDR gives a single scalar estimate of such differentials where further rankings can be made.

With this assessment, we could notice that the current state-of-the-art Face Recognition systems based on ArcFace loss are fairer than the state-of-the-art from a few years ago and with the evaluated COTS system.
Worth noting that the DCNNs based on ArcFace do not have any fairness constraints.
We can hypothesize that the margins imposed in the ArcFace loss play a role in minimizing within-class variability and maximizing between-class separation (embeddings from the same identity tends to be more compact than DCNNs training ``vanilla'' cross-entropy loss).
This possibly impacted the scoring behavior allowing us to ``safely'' use fair decision thresholds.

We could observe in Table \ref{tb:morph} with the MORPH dataset that false alarms are more frequent with Asian, Hispanic, and Black subjects (using operation points around $10^{-2}$).
The same trends were observed with female subjects using the Mobio dataset.
We can also observe the number of false alarms using the comparison scores between non-homogeneous samples (gallery and probe samples from different demographics) is substantially lower than homogeneous samples (gallery and probe samples from different demographic groups), which corroborates with the findings of the literature\cite{grother2019face,howard2019effect,michalski2018impact}.

FDR can be trivially extended to other biometric recognition tasks, such as closed-set or opened-set identification.
For instance, for \textbf{closed-set} identification, the most common figure of merit to compare two biometric systems is the rank-n\cite[chap.21]{jain2011handbook} defined as:

\begin{equation}
  \text{rank-n}=\frac{|C(n)|}{|P_g|}
  \label{eq:rank}
\end{equation}
,where $C(n)$ is the cumulative count of the number of probes of rank `n` or less and $|P_{\mathbb{G}}|$ corresponds to the size of the gallery $\mathbb{G}$.
Then, eq. \ref{eq:preA} could be rewritten to assess demographic differentials in \textbf{closed-set} identification systems as:

\begin{equation}
 C(\tau) = \max(|\text{rank-n}^{d_i}- \text{rank-n}^{d_j}|) \text{ } \forall d_i, d_j \in \mathcal{D},
  \label{eq:rank_fair}
\end{equation}

In \textbf{opened-set} identification, a probe sample $p_j \in \mathbb{G}$ is correctly detected and identified if and only if the following conditions are met:
\begin{itemize}
 \item $\text{rank}-n(p_j)=1$
 \item $s*j \geq \tau$, where $s*j$ corresponds to the comparison score between $p_j$ and its unique match in $\mathbb{G}$, and $\tau$ is an operational threshold.
\end{itemize}
Then a figure of merit called Detection and Identification Rate $\text{DIR}(\tau)$ that measures the fractions of the probes that were correctly detected and identified in the gallery $\mathbb{G}$ \cite[chap.21,p.553]{jain2011handbook}  can be computed.
There is a second figure of merit computed in the opened-set identification problem, which is the False-Alarm Rate ($FAR(\tau)$) which measures the proportion of zeroth effort impostor probe samples $p_j \notin \mathbb{G}$ that was incorrectly matched in  $\mathbb{G}$.

These two figure merits can be similarly combined as in eq. \ref{eq:fairness} as: 
\begin{equation}
  \begin{aligned}
    & \text{FDR'}(\tau)= \alpha \max(|\text{FAR}(\tau)^{d_i}- \text{FAR}(\tau)^{d_j}|) + \\
    &  (1-\alpha)\max(|\text{DIR}(\tau)^{d_i}- \text{DIR}(\tau)^{d_j}|) \\
    & \forall d_i, d_j \in \mathcal{D},
   \end{aligned} 
  \label{eq:fair-open}
\end{equation}
,where alpha has the same role as described in section 3.2.

Further assessment of those extensions for identification will be carried out as future work.




\section{Conclusions}

In this work, we introduced the Fairness Discrepancy Rate (FDR) to assess demographic differentials in biometric verification systems.
FDR tackles a threshold problem, which is the main issue on how the biometric community address such differentials.
We argue that such a figure of merit is a proxy to the separation fairness criteria since we aim that different demographic groups experience the same error rates.
A substantial amount of works in the biometrics community assess demographic differentials in verification systems by comparing DET curves or ROC curves of different demographic groups separately.
This type of comparison assumes that decision thresholds are demographic-specific, which is not feasible or ethical in operational conditions.
FDR addresses that by assessing demographic differentials assuming single decision thresholds.
Fair biometric recognition systems are fair if a decision threshold $\tau$ is ``fair'' for all demographic groups concerning $\text{FMR}(\tau)$ and $\text{FNMR}(\tau)$ and FDR proxies this behavior.
Furthermore, the $\text{FMR}(\tau)$ and $\text{FNMR}(\tau)$ trade-off with respect to fair behavior can be set by addressing the value $\alpha$ in Equation \ref{eq:fairness}.
Finally, the Area Under FDR provides a general overview of demographic differentials under a range of decision thresholds and allows a quick comparison between different biometric verification systems.

Two groups of experiments were carried out to evaluate this new figure of merit.
In the first one, a case study using synthetic data was presented, and it was demonstrated how FDR behaves in extreme cases of fair and unfair scenarios.
In the second, a case study using four different face verification systems and three databases was carried out.
We could observe via the FDR plots that all evaluated face verification systems presents gender and racial biases to some degree. 
Furthermore, it was possible to quickly compare different face recognition systems concerning their demographic discrepancies using the Area Under FDR.
Worth noting that neither FDR nor Area Under FDR is direct proxies for how ``accurate'' a biometric verification system is.
Possible error rates have to be analyzed in parallel to have a full picture of the trade-off between accuracy vs. fairness.

We also briefly presented a possible trivial extension of such a figure of merit to closed and opened-set identification problems.
Further work in this direction will be carried out.

For reproducibility purposes of the work, all the source code, trained models, and recognition scores are made publicly available.

We hope that these tools are useful for the biometrics community to assess demographic differentials, and we advocate for some standardization towards ISO/IEC 19795-10\footnote{https://www.iso.org/standard/81223.html}.
Hence, demographic differential can be easily assessed as any other figure of merit, such as $\text{FMR}$ or $\text{FNMR}$.


\vspace{-1.5em}

\bibliographystyle{IEEEtran}
\bibliography{bibliography}


\begin{IEEEbiography}
 [{\includegraphics[width=1in,height=1.25in,clip,keepaspectratio]{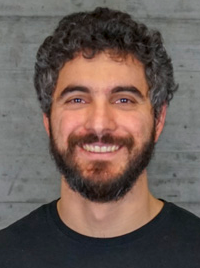}}]{Tiago de Freitas Pereira}
received the Ph.D. degree in Electrical Engineering from Ecole Polytechnique Federale de Lausanne ( http://www.epfl.ch) in Switzerland (2019) whose main contribution was the development of the Domain Specific Units (DSU) framework for Heterogeneous Face Recognition.
Currently he is a postdoctoral researcher at the Idiap Research Institute in the Biometrics Security and Privacy group and conducts research on face recognition.
\end{IEEEbiography}

\vspace{-50em}

\begin{IEEEbiography}
 [{\includegraphics[width=1in,height=1.25in,clip,keepaspectratio]{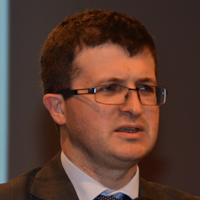}}]{S{\'e}bastien Marcel}
is the head of the Biometrics group and senior research scientist at the Idiap Research Institute (CH), where he leads a research team and conducts research on biometrics and more particularly face recognition, speaker recognition, vein recognition and spoofing/anti-spoofing.
He received the Ph.D. degree in signal processing from Universit{\'e} de Rennes I in France (2000) at CNET, the research center of France Telecom (now Orange Labs). He is currently interested in pattern recognition and machine learning with a focus on multimodal biometric person recognition.

\end{IEEEbiography}




\end{document}


\title{\textit{Supplementary Material: Fairness in Biometrics: a figure of merit to assess biometric verification systems}}

\author{Tiago de Freitas Pereira, \and S{\'e}bastien Marcel\\
        Idiap Research Institute \\
        \small{\{tiago.pereira,sebastien.marcel\}@idiap.ch}


}

\markboth{}{}

\IEEEtitleabstractindextext{%
\begin{abstract}

Machine learning-based (ML) systems are being largely deployed since the last decade in a myriad of scenarios impacting several instances in our daily lives.
With this vast sort of applications, aspects of fairness start to rise in the spotlight due to the social impact that this can get in some social groups.
In this work aspects of fairness in biometrics are addressed.
First, we introduce the first figure of merit that is able to evaluate and compare fairness aspects between multiple biometric verification systems, the so-called Fairness Discrepancy Rate (FDR).
A use case with two synthetic biometric systems is introduced and demonstrates the potential of this figure of merit in extreme cases of demographic differentials.
Second, a use case using face biometrics is presented where several systems are evaluated compared with this new figure of merit using three public datasets exploring gender and race demographics.

\end{abstract}


}

\maketitle

\IEEEdisplaynontitleabstractindextext

\IEEEpeerreviewmaketitle

\appendices

\begin{center}
\section{Face Recognition Use Case: Extra experiments}
\end{center}
\label{app:face_rec_use_case}

In this appendix extra experiments using the other three face verification baselines (COTS, Inception-Resnet-v2 and Resnet50) are presented.

\subsection{MEDS II database (Fairness with respect to race)}

Tables \ref{tb:meds_inception}, \ref{tb:meds_resnet50} and \ref{tb:meds_rankone} presents the $\text{FMR}(\tau)$, $\text{FNMR}(\tau)$ and $\text{FDR}(\tau)$ in the test set for the Inception-Resnet-v2, Resnet50 and the COTS.
It is possible to observe that Inception-Resnet-v2 and Resnet50 have similar trends in terms of FMR.
Both tend to raise more false alarms for Black subjects for operational points around $x \leq 10^{-2}$.
However, Inception-Resnet-v2 tends to falsely reject more White subjects from $x \geq 10^{-3}$ while Resnet50 presents fewer discrepancies between the two races when $x$ decreases.
The COTS system is the fairest one; however, as mentioned in the main paper, fairness is not necessarily a proxy for low FMR and FNMR.
This system presented the highest FNMR values no matter the decision threshold.
This is mostly because its scoring function is $[0,1]$ bounded and easily saturates around those bounds, making difficult the task of picking $\tau$.

\begin{table}[ht]
   \begin{center}

   \caption{MEDS II - Inception Resnet v2: $\text{FNMR}(\tau)$, $\text{FMR}(\tau)$, and $\text{FDR}(\tau)$ per demographic (Demog.) in the test set. These figures of merit are fragmented by the race of the samples used for enrollment and the race of the samples used for probe (``(e-p)'' in the table.)\textsuperscript{*}.}

    \begin{tabular}{|r||rrrrr|}
      \hline
  
       $\tau=FMR_{10^{x}}$  &  $10^{-1}$ &   $10^{-2}$ &   $10^{-3}$ &   $10^{-4}$ &   $10^{-5}$  \\
	
      \hline
      Demog (e-p) & \multicolumn{5}{c|}{$\text{FMR}_x(\tau)$} \\
      \hline
  White - White & \colorMyCellMEDS{0.095}  &  \colorMyCellMEDS{0.005} &  \colorMyCellMEDS{0.001}  &     \colorMyCellMEDS{0}    &    \colorMyCellMEDS{0} \\
  White - Black & \colorMyCellMEDS{0.029}  &  \colorMyCellMEDS{0     }   &  \colorMyCellMEDS{0     }    &     \colorMyCellMEDS{0}    &    \colorMyCellMEDS{0} \\
  Black - White & \colorMyCellMEDS{0.028}  &  \colorMyCellMEDS{0     }   &  \colorMyCellMEDS{0     }    &     \colorMyCellMEDS{0}    &    \colorMyCellMEDS{0} \\
  Black - Black & \colorMyCellMEDS{0.235}  & \colorMyCellMEDS{0.028}  & \colorMyCellMEDS{0.003}   &     \colorMyCellMEDS{0}    &    \colorMyCellMEDS{0} \\
      \hline
      & \multicolumn{5}{c|}{$\text{FNMR}_x(\tau)$} \\
      
      \hline
  White - White & \colorMyCellMEDS{0} &   \colorMyCellMEDS{0.016} &    \colorMyCellMEDS{0.082}  &   \colorMyCellMEDS{0.115}  &   \colorMyCell{0.148} \\
  Black - Black & \colorMyCellMEDS{0} &   \colorMyCellMEDS{0.007} &    \colorMyCellMEDS{0.020}  &    \colorMyCellMEDS{0.067}  & \colorMyCell{ 0.101} \\

       \hline
        $FDR(\tau)$ & 0.930  &  0.984  &   0.968  &  0.976 &  0.976 \\
       \hline        
      
   \end{tabular}

  \label{tb:meds_inception}
  \end{center}  
  
  \small\textsuperscript{\textsuperscript{*}In this example $\tau=\text{FMR}^{\text{dev}}_{x}$ where \textbf{dev} is the development-set.}
  \small\textsuperscript{$\text{FMR}(\tau)$, $\text{FNMR}(\tau)$ and $FDR(\tau)$ are reported using the test-set}
  
\end{table}

\begin{table}[ht]
   \begin{center}

   \caption{MEDS II - Resnet50: $\text{FNMR}(\tau)$, $\text{FMR}(\tau)$, and $\text{FDR}(\tau)$ per demographic (Demog.) in the test set. These figures of merit are fragmented by the race of the samples used for enrollment and the race of the samples used for probe (``(e-p)'' in the table.). In this example $\tau=\text{FMR}^{\text{dev}}_{x}$ where \textbf{dev} is the development set.}

    \begin{tabular}{|r||rrrrr|}
      \hline
  
     $\tau=FMR_{10^{-x}}$ &  $10^{-1}$ &   $10^{-2}$ &   $10^{-3}$ &   $10^{-4}$ &   $10^{-5}$  \\
	
      \hline
      Demog (e-p) & \multicolumn{5}{c|}{$\text{FMR}_x(\tau)$} \\
      \hline
  White - White & \colorMyCellMEDSgabor{0.13}  &  \colorMyCellMEDSgabor{0.008} &  \colorMyCellMEDSgabor{0.001} &     \colorMyCellMEDSgabor{0.0}        &    \colorMyCellMEDSgabor{0.0} \\
  White - Black & \colorMyCellMEDSgabor{0.05}  &  \colorMyCellMEDSgabor{0.003} &  \colorMyCellMEDSgabor{0.0}      &     \colorMyCellMEDSgabor{0.0}         &    \colorMyCellMEDSgabor{0.0} \\
  Black - White & \colorMyCellMEDSgabor{0.042}  &  \colorMyCellMEDSgabor{0.002}  &  \colorMyCellMEDSgabor{0.0}   &     \colorMyCellMEDSgabor{0.0}         &    \colorMyCellMEDSgabor{0.0} \\
  Black - Black & \colorMyCellMEDSgabor{0.152}  & \colorMyCellMEDSgabor{0.021}  &  \colorMyCellMEDSgabor{0.003} &     \colorMyCellMEDSgabor{0.001}    &    \colorMyCellMEDSgabor{0.0} \\
      \hline
      & \multicolumn{5}{c|}{$\text{FNMR}_x(\tau)$} \\
      
      \hline
  White - White & \colorMyCellMEDSgabor{0}        &   \colorMyCellMEDSgabor{0.016} &  \colorMyCellMEDSgabor{0.098}  & \colorMyCellMEDSgabor{0.131}  & \colorMyCellMEDSgabor{0.148} \\
  Black - Black & \colorMyCellMEDSgabor{0.007} &   \colorMyCellMEDSgabor{0.033} &  \colorMyCellMEDSgabor{0.073}  & \colorMyCellMEDSgabor{0.133}  & \colorMyCellMEDSgabor{0.16} \\

       \hline
        $FDR(\tau)$ & 0.985  & 0.985  & 0.986 &  0.998 &  0.994 \\
       \hline        
      
   \end{tabular}

  \label{tb:meds_resnet50}
  \end{center}  
\end{table}

\begin{table}[ht]
   \begin{center}

   \caption{MEDS II - RankOne: $\text{FNMR}(\tau)$, $\text{FMR}(\tau)$, and $\text{FDR}(\tau)$ per demographic (Demog.) in the test set. These figures of merit are fragmented by the race of the samples used for enrollment and the race of the samples used for probe (``(e-p)'' in the table.). In this example $\tau=\text{FMR}^{\text{dev}}_{x}$ where \textbf{dev} is the development set.}

    \begin{tabular}{|r||rrrrr|}
      \hline
  
      $\tau=FMR_{10^{-x}}$ &  $10^{-1}$ &   $10^{-2}$ &   $10^{-3}$ &   $10^{-4}$ &   $10^{-5}$  \\
	
      \hline
      Demog (e-p) & \multicolumn{5}{c|}{$\text{FMR}_x(\tau)$} \\
      \hline
  White - White & \colorMyCellMEDSrone{0.09}  &  \colorMyCellMEDSrone{0.007} &  \colorMyCellMEDSrone{0.0}   &     \colorMyCellMEDSrone{0.0}    &    \colorMyCellMEDSrone{1.0} \\
  White - Black & \colorMyCellMEDSrone{0.104} &  \colorMyCellMEDSrone{0.010} &  \colorMyCellMEDSrone{0.001} &     \colorMyCellMEDSrone{0.0}    &    \colorMyCellMEDSrone{1.0} \\
  Black - White & \colorMyCellMEDSrone{0.090} &  \colorMyCellMEDSrone{0.008} &  \colorMyCellMEDSrone{0.0}   &     \colorMyCellMEDSrone{0.0}    &    \colorMyCellMEDSrone{1.0} \\
  Black - Black & \colorMyCellMEDSrone{0.096} & \colorMyCellMEDSrone{0.008}  &  \colorMyCellMEDSrone{0.001} &     \colorMyCellMEDSrone{0.0}    &    \colorMyCellMEDSrone{1.0} \\
      \hline
      & \multicolumn{5}{c|}{$\text{FNMR}_x(\tau)$} \\
      
      \hline
  White - White & \colorMyCellMEDSrone{0.864}  &   \colorMyCellMEDSrone{1.0} &  \colorMyCellMEDSrone{1.0}  & \colorMyCellMEDSrone{1.0}  & \colorMyCellMEDSrone{1.0} \\
  Black - Black & \colorMyCellMEDSrone{0.953} &   \colorMyCellMEDSrone{1.0} &  \colorMyCellMEDSrone{1.0}  & \colorMyCellMEDSrone{1.0}  & \colorMyCellMEDSrone{1.0} \\

       \hline
        $FDR(\tau)$ & 0.952  &  0.995  &   0.999  &  1.0 &  1.0 \\
       \hline        
      
   \end{tabular}

  \label{tb:meds_rankone}
  \end{center}  
\end{table}

\newpage
\subsection{Mobio database (Fairness with respect to gender)}

Tables \ref{tb:mobio_inception}, \ref{tb:mobio_resnet}, and \ref{tb:mobio_rankone} presents the $\text{FMR}(\tau)$, $\text{FNMR}(\tau)$ and $\text{FDR}(\tau)$ in the test set for the Inception-Resnet-v2, Resnet50 and the COTS.
It is possible to observe that Inception-Resnet-v2 and Resnet50 have similar trends in terms of FMR and FNMR.
Both tend to raise more false alarms for female subjects for operational points around $x \leq 10^{-2}$.
And falsely reject more female subjects when $x \geq 10^{-4}$.
The COTS also falsely reject more female subjects for $x \leq 10^{-2}$ ($FDR(\tau=10^{-1})$), but it presents few disparities in terms of False Non-Matches when $x$ decreases, and this can be spotted in terms of FDR ($FDR \approx 0.99$ for $x \geq 10^{-3}$).

\begin{table}[ht]
   \begin{center}

   \caption{MOBIO - Inception Resnet v2: $\text{FNMR}(\tau)$, $\text{FMR}(\tau)$, and $\text{FDR}(\tau)$ per gender in the test set. These figures of merit are fragmented by the gender of the samples used for enrollment and the race of the samples used for probe (``(e-p)'' in the table.)\textsuperscript{*}.}

    \begin{tabular}{|r||rrrrr|}
      \hline
  
      $\tau=FMR_{10^{x}}$  &  $10^{-1}$ &   $10^{-2}$ &   $10^{-3}$ &   $10^{-4}$ &   $10^{-5}$  \\
	
      \hline
      Demog.(e-p) & \multicolumn{5}{c|}{$\text{FMR}_x(\tau)$} \\
      \hline
  Male-Male   & \colorMyCellMOBIO{0.067}  & \colorMyCellMOBIO{0.002}  &   \colorMyCellMOBIO{0}     & \colorMyCellMOBIO{0}       & \colorMyCellMOBIO{0}\\
  Male-Female & \colorMyCellMOBIO{0.014}  & \colorMyCellMOBIO{0}      &   \colorMyCellMOBIO{0}     & \colorMyCellMOBIO{0}       & \colorMyCellMOBIO{0}  \\
  Female-Male  & \colorMyCellMOBIO{0.011} & \colorMyCellMOBIO{0}      &   \colorMyCellMOBIO{0}     &  \colorMyCellMOBIO{0}      & \colorMyCellMOBIO{0} \\
  Female-Female & \colorMyCellMOBIO{0.28}  & \colorMyCellMOBIO{0.029} &   \colorMyCellMOBIO{0.004} &  \colorMyCellMOBIO{0.001}  & \colorMyCellMOBIO{0}\\
      \hline
      & \multicolumn{5}{c|}{$\text{FNMR}_x(\tau)$} \\
      
      \hline
  Male-Male     & \colorMyCellMOBIO{0.001} & \colorMyCellMOBIO{0.003} &  \colorMyCellMOBIO{0.015} & \colorMyCellMOBIO{0.039}  &  \colorMyCellMOBIO{0.106} \\
  Female-Female & \colorMyCellMOBIO{0}     &  \colorMyCellMOBIO{0.01} & \colorMyCellMOBIO{0.045}  & \colorMyCellMOBIO{0.09}   &  \colorMyCellMOBIO{0.21} \\

       \hline
        $FDR(\tau)$ & 0.893 &  0.983  &  0.983  &   0.974  &  0.948 \\
       \hline        
      
   \end{tabular}

  \label{tb:mobio_inception}
  \end{center}
    \small\textsuperscript{\textsuperscript{*}In this example $\tau=\text{FMR}^{\text{dev}_{x}}$ where \textbf{dev} is the development-set.}
  \small\textsuperscript{$\text{FMR}(\tau)$, $\text{FNMR}(\tau)$ and $FDR(\tau)$ are reported using the test-set}

\end{table}

\begin{table}[ht]
   \begin{center}

   \caption{MOBIO - Resnet50: $\text{FNMR}(\tau)$, $\text{FMR}(\tau)$, and $\text{FDR}(\tau)$ per gender in the test set. These figures of merit are fragmented by the gender of the samples used for enrollment and the race of the samples used for probe (``(e-p)'' in the table.)\textsuperscript{*}.}

    \begin{tabular}{|r||rrrrr|}
      \hline
  
      $\tau=FMR_{10^{x}}$  &  $10^{-1}$ &   $10^{-2}$ &   $10^{-3}$ &   $10^{-4}$ &   $10^{-5}$  \\
	
      \hline
      Demog.(e-p) & \multicolumn{5}{c|}{$\text{FMR}_x(\tau)$} \\
      \hline
  Male-Male   & \colorMyCellMOBIOResnet{0.072}  & \colorMyCellMOBIOResnet{0.004}   &   \colorMyCellMOBIOResnet{0}     & \colorMyCellMOBIOResnet{0}       & \colorMyCellMOBIOResnet{0}\\
  Male-Female & \colorMyCellMOBIOResnet{0.049}  & \colorMyCellMOBIOResnet{0.001}   &   \colorMyCellMOBIOResnet{0}     & \colorMyCellMOBIOResnet{0}       & \colorMyCellMOBIOResnet{0}  \\
  Female-Male  & \colorMyCellMOBIOResnet{0.034} & \colorMyCellMOBIOResnet{0.001}   &   \colorMyCellMOBIOResnet{0}     &  \colorMyCellMOBIOResnet{0}      & \colorMyCellMOBIOResnet{0} \\
  Female-Female & \colorMyCellMOBIOResnet{0.304}  & \colorMyCellMOBIOResnet{0.062} &   \colorMyCellMOBIOResnet{0.004} &  \colorMyCellMOBIOResnet{0.001}  & \colorMyCellMOBIOResnet{0}\\
      \hline
      & \multicolumn{5}{c|}{$\text{FNMR}_x(\tau)$} \\
      
      \hline
  Male-Male     & \colorMyCellMOBIOResnet{0.002} & \colorMyCellMOBIOResnet{0.011} & \colorMyCellMOBIOResnet{0.073} & \colorMyCellMOBIOResnet{0.224}  &  \colorMyCellMOBIOResnet{0.441} \\
  Female-Female & \colorMyCellMOBIOResnet{0.001} & \colorMyCellMOBIOResnet{0.016} & \colorMyCellMOBIOResnet{0.077} & \colorMyCellMOBIOResnet{0.278}  &  \colorMyCellMOBIOResnet{0.561} \\

       \hline
        $FDR(\tau)$ & 0.883 & 0.968  &  0.996  &   0.973  &   0.94 \\
       \hline        
      
   \end{tabular}

  \label{tb:mobio_resnet}
  \end{center}
    \small\textsuperscript{\textsuperscript{*}In this example $\tau=\text{FMR}^{\text{dev}_{x}}$ where \textbf{dev} is the development-set.}
  \small\textsuperscript{$\text{FMR}(\tau)$, $\text{FNMR}(\tau)$ and $FDR(\tau)$ are reported using the test-set}

\end{table}

\begin{table}[ht]
   \begin{center}

   \caption{MOBIO - RankOne: $\text{FNMR}(\tau)$, $\text{FMR}(\tau)$, and $\text{FDR}(\tau)$ per gender in the test set. These figures of merit are fragmented by the gender of the samples used for enrollment and the race of the samples used for probe (``(e-p)'' in the table.). In this example $\tau=\text{FMR}^{\text{dev}}_{x}$ where \textbf{dev} is the development set.}

    \begin{tabular}{|r||rrrrr|}
      \hline
  
      $\tau=FMR_{10^{-x}}$ &  $10^{-1}$ &   $10^{-2}$ &   $10^{-3}$ &   $10^{-4}$ &   $10^{-5}$ \\
	
      \hline
      Demog.(e-p) & \multicolumn{5}{c|}{$\text{FMR}_x(\tau)$} \\
      \hline
  Male-Male     & \colorMyCellMOBIOrankone{0.073} & \colorMyCellMOBIOrankone{0.005} & \colorMyCellMOBIOrankone{0.0}   & \colorMyCellMOBIOrankone{0.0}  & \colorMyCellMOBIOrankone{0.0}\\
  Male-Female   & \colorMyCellMOBIOrankone{0.048} & \colorMyCellMOBIOrankone{0.002} & \colorMyCellMOBIOrankone{0.0}   & \colorMyCellMOBIOrankone{0.0}  & \colorMyCellMOBIOrankone{0.0}  \\
  Female-Male   & \colorMyCellMOBIOrankone{0.049} & \colorMyCellMOBIOrankone{0.002} & \colorMyCellMOBIOrankone{0.0}   & \colorMyCellMOBIOrankone{0.0}  & \colorMyCellMOBIOrankone{0.0} \\
  Female-Female & \colorMyCellMOBIOrankone{0.207} & \colorMyCellMOBIOrankone{0.028} & \colorMyCellMOBIOrankone{0.003} & \colorMyCellMOBIOrankone{0.0}  & \colorMyCellMOBIOrankone{0.0}  \\
      \hline
      & \multicolumn{5}{c|}{$\text{FNMR}_x(\tau)$} \\
      
      \hline
  Male-Male     & \colorMyCellMOBIOrankone{0.0} & \colorMyCellMOBIOrankone{0.001} & \colorMyCellMOBIOrankone{0.001} & \colorMyCellMOBIOrankone{0.005}  &  \colorMyCellMOBIOrankone{0.011} \\
  Female-Female & \colorMyCellMOBIOrankone{0.0} & \colorMyCellMOBIOrankone{0.0}   & \colorMyCellMOBIOrankone{0.002} & \colorMyCellMOBIOrankone{0.011}  &  \colorMyCellMOBIOrankone{0.018} \\

       \hline
        $FDR(\tau)$ & 0.933 & 0.988  & 0.998 &  0.997 & 0.9965 \\
       \hline        
      
   \end{tabular}

  \label{tb:mobio_rankone}
  \end{center}  
\end{table}

\newpage
\subsection{Morph database (Fairness with respect to race)}

This section fairness with respect to race is assessed using FDR using the MORPH dataset
This section is split in two subsections, the first one we continue our analysis from Section 4.4 using the male subjects.
Then, the following subsection analyses the racial effects in female subjects. 

\subsubsection{Male Cohort}

Tables \ref{tb:morph}, \ref{tb:morph_resenet}, and \ref{tb:morph_rankone} presents, respectively, the $\text{FMR}(\tau)$, $\text{FNMR}(\tau)$ and $\text{FDR}(\tau)$ in the test set for the Inception-Resnet-v2, Resnet50 and the COTS verification systems.
It can be observed that all verification systems tend to raise more false-alarms for Asian and Hispanic subjects.
This makes FDR substantially decrease to below $0.9$
In terms of false-non-matches, the COTS system presents less FNMR for the range of analyzed decision thresholds.
The impact of that is that FDR increases once $x$ is decreases.
It is also possible to observe that Inception-Resnet-v2 and Resnet50 systems tend to falsely non-match more White subjects once $x$ decreases.

\begin{table*}[ht]
   \begin{center}

   \caption{MORPH - Inception-Resnet-v2: $\text{FNMR}(\tau)$, $\text{FMR}(\tau)$, and $\text{FDR}(\tau)$ per demographic (Demog.) in the test set. These figures of merit are fragmented by the race of the samples used for enrollment and the race of the samples used for probe (``(e-p)'' in the table.)\textsuperscript{*}}

\begin{adjustwidth}{-3.5em}{}

    \begin{tabular}{|r||cccccc|}
      \hline
  
       $\tau=FMR_{10^{x}}$  &  $10^{-1}$ &   $10^{-2}$ &   $10^{-3}$ &   $10^{-4}$ &   $10^{-5}$ &   $10^{-6}$ \\
	
      \hline
      Demog. (e-p) & \multicolumn{6}{c|}{$\text{FMR}_x(\tau)$} \\
      \hline
Asian - Asian & \colorMyCellMORPH{0.587}  &  \colorMyCellMORPH{0.231}  &  \colorMyCellMORPH{0.017} &   \colorMyCellMORPH{0}     & \colorMyCellMORPH{0}   &  \colorMyCellMORPH{0}  \\
Asian - Black & \colorMyCellMORPH{0.018}  &  \colorMyCellMORPH{0}      &  \colorMyCellMORPH{0}     &    \colorMyCellMORPH{0}    & \colorMyCellMORPH{0}   &  \colorMyCellMORPH{0}  \\
Asian - Hisp. & \colorMyCellMORPH{0.113}  &  \colorMyCellMORPH{0.004}  &  \colorMyCellMORPH{0}     &   \colorMyCellMORPH{0}     & \colorMyCellMORPH{0}   &  \colorMyCellMORPH{0}  \\
Asian - White & \colorMyCellMORPH{0.007}  &  \colorMyCellMORPH{0}      &  \colorMyCellMORPH{0}     &   \colorMyCellMORPH{0}     & \colorMyCellMORPH{0}   &  \colorMyCellMORPH{0}  \\
Black - Asian & \colorMyCellMORPH{0.025}  &  \colorMyCellMORPH{0}      &  \colorMyCellMORPH{0}     &   \colorMyCellMORPH{0}     & \colorMyCellMORPH{0}   &  \colorMyCellMORPH{0}  \\
Black - Black & \colorMyCellMORPH{0.184}  &  \colorMyCellMORPH{0.017}  &  \colorMyCellMORPH{0.001} &   \colorMyCellMORPH{0}     & \colorMyCellMORPH{0}   &  \colorMyCellMORPH{0}  \\
Black - Hisp. & \colorMyCellMORPH{0.020}  &  \colorMyCellMORPH{0}      &  \colorMyCellMORPH{0}     &   \colorMyCellMORPH{0}     & \colorMyCellMORPH{0}   &  \colorMyCellMORPH{0}    \\
Black - White & \colorMyCellMORPH{0.007}  &  \colorMyCellMORPH{0}      &  \colorMyCellMORPH{0}     &   \colorMyCellMORPH{0}     & \colorMyCellMORPH{0}   &  \colorMyCellMORPH{0}  \\
Hisp. - Asian & \colorMyCellMORPH{0.12 }  &  \colorMyCellMORPH{0.006}  &  \colorMyCellMORPH{0.001} &   \colorMyCellMORPH{0.001} & \colorMyCellMORPH{0}   &  \colorMyCellMORPH{0}  \\
Hisp. - Black & \colorMyCellMORPH{0.017}  &  \colorMyCellMORPH{0}      &  \colorMyCellMORPH{0}     &   \colorMyCellMORPH{0}     & \colorMyCellMORPH{0}   &  \colorMyCellMORPH{0}  \\
Hisp. - Hisp. & \colorMyCellMORPH{0.202}  &  \colorMyCellMORPH{0.018}  &  \colorMyCellMORPH{0.001} &   \colorMyCellMORPH{0}     & \colorMyCellMORPH{0}   &  \colorMyCellMORPH{0}  \\
Hisp. - White & \colorMyCellMORPH{0.024}  & \colorMyCellMORPH{0.001}   &  \colorMyCellMORPH{0}     &   \colorMyCellMORPH{0}     & \colorMyCellMORPH{0}   &  \colorMyCellMORPH{0}  \\
White - Asian & \colorMyCellMORPH{0.01 }  &  \colorMyCellMORPH{0}      &  \colorMyCellMORPH{0}     &   \colorMyCellMORPH{0}     & \colorMyCellMORPH{0}   &  \colorMyCellMORPH{0}  \\
White - Black & \colorMyCellMORPH{0.007}  &  \colorMyCellMORPH{0}      &  \colorMyCellMORPH{0}     &   \colorMyCellMORPH{0}     & \colorMyCellMORPH{0}   &  \colorMyCellMORPH{0}  \\
White - Hisp. & \colorMyCellMORPH{0.024}  &  \colorMyCellMORPH{0.001}  &  \colorMyCellMORPH{0}     &   \colorMyCellMORPH{0}     & \colorMyCellMORPH{0}   &  \colorMyCellMORPH{0}  \\
White - White & \colorMyCellMORPH{0.066}  &  \colorMyCellMORPH{0.003}  &  \colorMyCellMORPH{0}     &   \colorMyCellMORPH{0}     & \colorMyCellMORPH{0}   &  \colorMyCellMORPH{0}  \\

      \hline
      & \multicolumn{6}{c|}{$\text{FNMR}_x(\tau)$} \\
      
      \hline

Asian - Asian & \colorMyCellMORPH{0}  & \colorMyCellMORPH{0}      & \colorMyCellMORPH{0}       &  \colorMyCellMORPH{0}       & \colorMyCellMORPH{0}      &    \colorMyCellMORPH{0}      \\
Black - Black & \colorMyCellMORPH{0}  & \colorMyCellMORPH{0.001}  & \colorMyCellMORPH{0.002}   &  \colorMyCellMORPH{0.01 }   & \colorMyCellMORPH{0.038}  &   \colorMyCellMORPH{0.106}  \\
Hisp. - Hisp. & \colorMyCellMORPH{0}  & \colorMyCellMORPH{0}      & \colorMyCellMORPH{0}       &  \colorMyCellMORPH{0.025}   & \colorMyCellMORPH{0.076}  &    \colorMyCellMORPH{0.19}  \\
White - White & \colorMyCellMORPH{0}  & \colorMyCellMORPH{0.002}  & \colorMyCellMORPH{0.015}   &  \colorMyCellMORPH{0.074}   & \colorMyCellMORPH{0.202}  &    \colorMyCellMORPH{0.355}  \\

       \hline
        $FDR(\tau)$ & 0.7395  & 0.885  &  0.984  &   0.963  &  0.899 &  0.8225 \\
       \hline        
      
   \end{tabular}

\end{adjustwidth}

  \label{tb:morph}
  \end{center}  
  
  \small\textsuperscript{\textsuperscript{*}In this example $\tau=\text{FMR}^{\text{dev}}_{x}$ where \textbf{dev} is the development-set.}
  \small\textsuperscript{$\text{FMR}(\tau)$, $\text{FNMR}(\tau)$ and $FDR(\tau)$ are reported using the test-set}
  
\end{table*}

\begin{table*}[ht]
   \begin{center}

   \caption{MORPH - Resnet50: $\text{FNMR}(\tau)$, $\text{FMR}(\tau)$, and $\text{FDR}(\tau)$ per demographic (Demog.) in the test set. These figures of merit are fragmented by the race of the samples used for enrollment and the race of the samples used for probe (``(e-p)'' in the table.). In this example $\tau=\text{FMR}^{\text{dev}}_{x}$ where \textbf{dev} is the development set.}

    \begin{tabular}{|r||cccccc|}
      \hline
  
      $\tau=FMR_{10^{-x}}$ &  $10^{-1}$ &   $10^{-2}$ &   $10^{-3}$ &   $10^{-4}$ &   $10^{-5}$ &   $10^{-6}$ \\
	
      \hline
      Demog. (e-p) & \multicolumn{6}{c|}{$\text{FMR}_x(\tau)$} \\
      \hline
Asian - Asian & \colorMyCellMORPHgabor{0.372}  &  \colorMyCellMORPHgabor{0.091} &  \colorMyCellMORPHgabor{0.0}    &   \colorMyCellMORPHgabor{0.0}   & \colorMyCellMORPHgabor{0.0}   &  \colorMyCellMORPHgabor{0.0}  \\
Asian - Black & \colorMyCellMORPHgabor{0.045}  &  \colorMyCellMORPHgabor{0.001} &  \colorMyCellMORPHgabor{0.0}    &    \colorMyCellMORPHgabor{0.0}  & \colorMyCellMORPHgabor{0.0}   &  \colorMyCellMORPHgabor{0.0}  \\
Asian - Hisp. & \colorMyCellMORPHgabor{0.146}  &  \colorMyCellMORPHgabor{0.015} &  \colorMyCellMORPHgabor{0.0}     &   \colorMyCellMORPHgabor{0.0}   & \colorMyCellMORPHgabor{0.0}   &  \colorMyCellMORPHgabor{0.0}  \\
Asian - White & \colorMyCellMORPHgabor{0.046}  &  \colorMyCellMORPHgabor{0.002} &  \colorMyCellMORPHgabor{0.0}    &   \colorMyCellMORPHgabor{0.0}   & \colorMyCellMORPHgabor{0.0}   &  \colorMyCellMORPHgabor{0.0}  \\
Black - Asian & \colorMyCellMORPHgabor{0.051}  &  \colorMyCellMORPHgabor{0.002} &  \colorMyCellMORPHgabor{0.0}     &   \colorMyCellMORPHgabor{0.0}   & \colorMyCellMORPHgabor{0.0}   &  \colorMyCellMORPHgabor{0.0}  \\
Black - Black & \colorMyCellMORPHgabor{0.158}  &  \colorMyCellMORPHgabor{0.018} &  \colorMyCellMORPHgabor{0.002} &   \colorMyCellMORPHgabor{0.0}   & \colorMyCellMORPHgabor{0.0}   &  \colorMyCellMORPHgabor{0.0}  \\
Black - Hisp. & \colorMyCellMORPHgabor{0.054}  &  \colorMyCellMORPHgabor{0.003} &  \colorMyCellMORPHgabor{0.0}      &   \colorMyCellMORPHgabor{0.0}   & \colorMyCellMORPHgabor{0.0}   &  \colorMyCellMORPHgabor{0.0}    \\
Black - White & \colorMyCellMORPHgabor{0.039}  &  \colorMyCellMORPHgabor{0.001}  &  \colorMyCellMORPHgabor{0.0}    &   \colorMyCellMORPHgabor{0.0}   & \colorMyCellMORPHgabor{0.0}   &  \colorMyCellMORPHgabor{0.0}  \\
Hisp. - Asian & \colorMyCellMORPHgabor{0.177}  &  \colorMyCellMORPHgabor{0.015} &  \colorMyCellMORPHgabor{0.002}     &   \colorMyCellMORPHgabor{0.0}   & \colorMyCellMORPHgabor{0.0}   &  \colorMyCellMORPHgabor{0.0}  \\
Hisp. - Black & \colorMyCellMORPHgabor{0.049}  &  \colorMyCellMORPHgabor{0.002} &  \colorMyCellMORPHgabor{0.0}      &   \colorMyCellMORPHgabor{0.0}   & \colorMyCellMORPHgabor{0.0}   &  \colorMyCellMORPHgabor{0.0}  \\
Hisp. - Hisp. & \colorMyCellMORPHgabor{0.23}  &  \colorMyCellMORPHgabor{0.035}  &  \colorMyCellMORPHgabor{0.005}        &   \colorMyCellMORPHgabor{0.0} & \colorMyCellMORPHgabor{0.0}   &  \colorMyCellMORPHgabor{0.0}  \\
Hisp. - White & \colorMyCellMORPHgabor{0.088}  & \colorMyCellMORPHgabor{0.006}  &  \colorMyCellMORPHgabor{0.0}     &   \colorMyCellMORPHgabor{0.0}   & \colorMyCellMORPHgabor{0.0}   &  \colorMyCellMORPHgabor{0.0}  \\
White - Asian & \colorMyCellMORPHgabor{0.049}  &  \colorMyCellMORPHgabor{0.002} &  \colorMyCellMORPHgabor{0.0}    &   \colorMyCellMORPHgabor{0.0}   & \colorMyCellMORPHgabor{0.0}   &  \colorMyCellMORPHgabor{0.0}  \\
White - Black & \colorMyCellMORPHgabor{0.038}  &  \colorMyCellMORPHgabor{0.001}  &  \colorMyCellMORPHgabor{0.0}   &   \colorMyCellMORPHgabor{0.0}   & \colorMyCellMORPHgabor{0.0}   &  \colorMyCellMORPHgabor{0.0}  \\
White - Hisp. & \colorMyCellMORPHgabor{0.087}  &  \colorMyCellMORPHgabor{0.006}  &  \colorMyCellMORPHgabor{0.001}    &   \colorMyCellMORPHgabor{0.0}   & \colorMyCellMORPHgabor{0.0}   &  \colorMyCellMORPHgabor{0.0}  \\
White - White & \colorMyCellMORPHgabor{0.143}  &  \colorMyCellMORPHgabor{0.015} &  \colorMyCellMORPHgabor{0.001}   &   \colorMyCellMORPHgabor{0.0}   & \colorMyCellMORPHgabor{0.0}   &  \colorMyCellMORPHgabor{0.0}  \\

      \hline
      & \multicolumn{6}{c|}{$\text{FNMR}_x(\tau)$} \\
      
      \hline

Asian - Asian & \colorMyCellMORPHgabor{0.0}    & \colorMyCellMORPHgabor{0.0}      & \colorMyCellMORPHgabor{0.0}       &  \colorMyCellMORPHgabor{0.0}      & \colorMyCellMORPHgabor{0.091}  &    \colorMyCellMORPHgabor{0.182}      \\
Black - Black & \colorMyCellMORPHgabor{0.0}     & \colorMyCellMORPHgabor{0.001}  & \colorMyCellMORPHgabor{0.006}  &  \colorMyCellMORPHgabor{0.023} & \colorMyCellMORPHgabor{0.055}  &   \colorMyCellMORPHgabor{0.113}  \\
Hisp. - Hisp. & \colorMyCellMORPHgabor{0.006}  & \colorMyCellMORPHgabor{0.006}  & \colorMyCellMORPHgabor{0.006}   &  \colorMyCellMORPHgabor{0.012} & \colorMyCellMORPHgabor{0.075}  &    \colorMyCellMORPHgabor{0.139}  \\
White - White & \colorMyCellMORPHgabor{0.0}     & \colorMyCellMORPHgabor{0.004} & \colorMyCellMORPHgabor{0.016}   &  \colorMyCellMORPHgabor{0.048} & \colorMyCellMORPHgabor{0.145}  &    \colorMyCellMORPHgabor{0.258}  \\

       \hline
        $FDR(\tau)$ & 0.882 &  0.959 &  0.989  &   0.976  &  0.955  & 0.9275 \\
       \hline        
      
   \end{tabular}

  \label{tb:morph_resenet}
  \end{center}  
\end{table*}

\begin{table*}[ht]
   \begin{center}

   \caption{MORPH - COTS: $\text{FNMR}(\tau)$, $\text{FMR}(\tau)$, and $\text{FDR}(\tau)$ per demographic (Demog.) in the test set. These figures of merit are fragmented by the race of the samples used for enrollment and the race of the samples used for probe (``(e-p)'' in the table.). In this example $\tau=\text{FMR}^{\text{dev}}_{x}$ where \textbf{dev} is the development set.}

    \begin{tabular}{|r||cccccc|}
      \hline
  
     $\tau=FMR_{10^{-x}}$&  $10^{-1}$ &   $10^{-2}$ &   $10^{-3}$ &   $10^{-4}$ &   $10^{-5}$ &   $10^{-6}$ \\
	
      \hline
      Demog. (e-p) & \multicolumn{6}{c|}{$\text{FMR}_x(\tau)$} \\
      \hline
Asian - Asian & \colorMyCellMORPHrone{0.733}  &  \colorMyCellMORPHrone{0.2}   &  \colorMyCellMORPHrone{0.1}   &   \colorMyCellMORPHrone{0.0}   & \colorMyCellMORPHrone{0.0}   &  \colorMyCellMORPHrone{0.0}  \\
Asian - Black & \colorMyCellMORPHrone{0.07}   &  \colorMyCellMORPHrone{0.003} &  \colorMyCellMORPHrone{0.0}   &   \colorMyCellMORPHrone{0.0}  & \colorMyCellMORPHrone{0.0}   &  \colorMyCellMORPHrone{0.0}  \\
Asian - Hisp. & \colorMyCellMORPHrone{0.152}  &  \colorMyCellMORPHrone{0.045} &  \colorMyCellMORPHrone{0.015} &   \colorMyCellMORPHrone{0.0}   & \colorMyCellMORPHrone{0.0}   &  \colorMyCellMORPHrone{0.0}  \\
Asian - White & \colorMyCellMORPHrone{0.128}  &  \colorMyCellMORPHrone{0.008} &  \colorMyCellMORPHrone{0.0}   &   \colorMyCellMORPHrone{0.0}   & \colorMyCellMORPHrone{0.0}   &  \colorMyCellMORPHrone{0.0}  \\
Black - Asian & \colorMyCellMORPHrone{0.066}  &  \colorMyCellMORPHrone{0.003} &  \colorMyCellMORPHrone{0.0}   &   \colorMyCellMORPHrone{0.0}   & \colorMyCellMORPHrone{0.0}   &  \colorMyCellMORPHrone{0.0}  \\
Black - Black & \colorMyCellMORPHrone{0.111}  &  \colorMyCellMORPHrone{0.013} &  \colorMyCellMORPHrone{0.001} &   \colorMyCellMORPHrone{0.0}   & \colorMyCellMORPHrone{0.0}   &  \colorMyCellMORPHrone{0.0}  \\
Black - Hisp. & \colorMyCellMORPHrone{0.076}  &  \colorMyCellMORPHrone{0.006} &  \colorMyCellMORPHrone{0.0}   &   \colorMyCellMORPHrone{0.0}   & \colorMyCellMORPHrone{0.0}   &  \colorMyCellMORPHrone{0.0}    \\
Black - White & \colorMyCellMORPHrone{0.067}  &  \colorMyCellMORPHrone{0.003}  &  \colorMyCellMORPHrone{0.0}  &   \colorMyCellMORPHrone{0.0}   & \colorMyCellMORPHrone{0.0}   &  \colorMyCellMORPHrone{0.0}  \\
Hisp. - Asian & \colorMyCellMORPHrone{0.186}  &  \colorMyCellMORPHrone{0.025} &  \colorMyCellMORPHrone{0.004} &   \colorMyCellMORPHrone{0.0}   & \colorMyCellMORPHrone{0.0}   &  \colorMyCellMORPHrone{0.0}  \\
Hisp. - Black & \colorMyCellMORPHrone{0.073}  &  \colorMyCellMORPHrone{0.004} &  \colorMyCellMORPHrone{0.0}   &   \colorMyCellMORPHrone{0.0}   & \colorMyCellMORPHrone{0.0}   &  \colorMyCellMORPHrone{0.0}  \\
Hisp. - Hisp. & \colorMyCellMORPHrone{0.282}  &  \colorMyCellMORPHrone{0.055}  &  \colorMyCellMORPHrone{0.010}&   \colorMyCellMORPHrone{0.003} & \colorMyCellMORPHrone{0.0}   &  \colorMyCellMORPHrone{0.0}  \\
Hisp. - White & \colorMyCellMORPHrone{0.136}  & \colorMyCellMORPHrone{0.013}  &  \colorMyCellMORPHrone{0.001} &   \colorMyCellMORPHrone{0.0}   & \colorMyCellMORPHrone{0.0}   &  \colorMyCellMORPHrone{0.0}  \\
White - Asian & \colorMyCellMORPHrone{0.118}  &  \colorMyCellMORPHrone{0.009} &  \colorMyCellMORPHrone{0.0}   &   \colorMyCellMORPHrone{0.0}   & \colorMyCellMORPHrone{0.0}   &  \colorMyCellMORPHrone{0.0}  \\
White - Black & \colorMyCellMORPHrone{0.067}  &  \colorMyCellMORPHrone{0.003}  &  \colorMyCellMORPHrone{0.0}  &   \colorMyCellMORPHrone{0.0}   & \colorMyCellMORPHrone{0.0}   &  \colorMyCellMORPHrone{0.0}  \\
White - Hisp. & \colorMyCellMORPHrone{0.128}  &  \colorMyCellMORPHrone{0.010}  &  \colorMyCellMORPHrone{0.0}  &   \colorMyCellMORPHrone{0.0}   & \colorMyCellMORPHrone{0.0}   &  \colorMyCellMORPHrone{0.0}  \\
White - White & \colorMyCellMORPHrone{0.218}  &  \colorMyCellMORPHrone{0.033} &  \colorMyCellMORPHrone{0.001} &   \colorMyCellMORPHrone{0.0}   & \colorMyCellMORPHrone{0.0}   &  \colorMyCellMORPHrone{0.0}  \\

      \hline
      & \multicolumn{6}{c|}{$\text{FNMR}_x(\tau)$} \\
      
      \hline

Asian - Asian & \colorMyCellMORPHrone{0.0}  & \colorMyCellMORPHrone{0.0}  & \colorMyCellMORPHrone{0.0} &  \colorMyCellMORPHrone{0.0} & \colorMyCellMORPHrone{0.0}  &    \colorMyCellMORPHrone{0.0}  \\
Black - Black & \colorMyCellMORPHrone{0.0}  & \colorMyCellMORPHrone{0.0}  & \colorMyCellMORPHrone{0.0} &  \colorMyCellMORPHrone{0.0} & \colorMyCellMORPHrone{0.007}  &   \colorMyCellMORPHrone{0.007}  \\
Hisp. - Hisp. & \colorMyCellMORPHrone{0.0}  & \colorMyCellMORPHrone{0.0}  & \colorMyCellMORPHrone{0.0} &  \colorMyCellMORPHrone{0.0} & \colorMyCellMORPHrone{0.0}  &    \colorMyCellMORPHrone{0.0}  \\
White - White & \colorMyCellMORPHrone{0.0}  & \colorMyCellMORPHrone{0.0}  & \colorMyCellMORPHrone{0.0} &  \colorMyCellMORPHrone{0.0} & \colorMyCellMORPHrone{0.0}  &    \colorMyCellMORPHrone{0.0}  \\

       \hline
        $FDR(\tau)$ & 0.68  & 0.90  &  0.95  &   0.99  &  0.99 &  0.99 \\
       \hline        
      
   \end{tabular}

  \label{tb:morph_rankone}
  \end{center}  
\end{table*}

\subsubsection{Female Cohort}

Tables \ref{tb:morph_female_inception}, \ref{tb:morph_female_insight}, \ref{tb:morph_female_resnet}, and \ref{tb:morph_female_rank_one} presents the $\text{FMR}(\tau)$, $\text{FNMR}(\tau)$ and $\text{FDR}(\tau)$ in the test set for, respectively, the Inception-Resnet-v2, ArcFace-InsightFace, Resnet50 and the COTS verification systems.
Worth noting that the number of Hispanic and Asian female subjects available to run a biometric experiment in this dataset is very little (less than 10); hence, we exclude these two demographics from our analysis.
Hence, an analysis of the two factors (gender and race) cannot be made.
In this experiment, the range of decision thresholds is the same as from the experiments with male subjects; only the analysis of $\text{FMR}(\tau)$, $\text{FNMR}(\tau)$ and $\text{FDR}(\tau)$ are split by gender.
This represents real operational conditions.

We can observe that all DCNN based FR systems tend to falsely match more black female subjects than white for $x \leq 10^{-2}$.
This effect cannot be observed for the COTS that tend to falsely match more white female subjects.
We can hypothesize that datasets, where those FR systems were trained, might cause this effect.
Datasets scraped from the web nowadays represent the demographic of public figures, and the number of black women in these datasets is substantially lower than white men, for instance.
However, further study about this possible causality link needs to be carried out.
On the other side of the spectrum, all the DCNN based FR systems tend to falsely non-match more white female subjects than black for $x \geq 10^{-4}$.
The COTS present fewer discrepancies for this range of operational points.

\begin{table}[ht]
   \begin{center}

   \caption{MORPH-Female cohort - Inception-Resnet-v2: $\text{FNMR}(\tau)$, $\text{FMR}(\tau)$, and $\text{FDR}(\tau)$ per demographic (Demog.) in the test set. These figures of merit are fragmented by the race of the samples used for enrollment and the race of the samples used for probe (``(e-p)'' in the table.). In this example $\tau=\text{FMR}^{\text{dev}}_{x}$ where \textbf{dev} is the development set.}

    \begin{tabular}{|r||rrrrrr|}
      \hline
  
      $\tau=FMR_{10^{-x}}$ &  $10^{-1}$ &   $10^{-2}$ &   $10^{-3}$ &   $10^{-4}$ &   $10^{-5}$ & $10^{-6}$ \\
	
      \hline
      Demog (e-p) & \multicolumn{6}{c|}{$\text{FMR}_x(\tau)$} \\
      \hline
  White - White & \colorMyCellMorphFemaleInception{0.187}  &  \colorMyCellMorphFemaleInception{0.014} &  \colorMyCellMorphFemaleInception{0.001}  &     \colorMyCellMorphFemaleInception{0.0}    &    \colorMyCellMorphFemaleInception{0.0} &  \colorMyCellMorphFemaleInception{0.0}\\
  White - Black & \colorMyCellMorphFemaleInception{0.019} &  \colorMyCellMorphFemaleInception{0.0}    &  \colorMyCellMorphFemaleInception{0.0}    &     \colorMyCellMorphFemaleInception{0.0}    &    \colorMyCellMorphFemaleInception{0.0} &  \colorMyCellMorphFemaleInception{0.0}\\
  Black - White & \colorMyCellMorphFemaleInception{0.020} &  \colorMyCellMorphFemaleInception{0.0}    &  \colorMyCellMorphFemaleInception{0.0}    &     \colorMyCellMorphFemaleInception{0.0}    &    \colorMyCellMorphFemaleInception{0.0} &  \colorMyCellMorphFemaleInception{0.0}\\
  Black - Black & \colorMyCellMorphFemaleInception{0.594} & \colorMyCellMorphFemaleInception{0.164}   &  \colorMyCellMorphFemaleInception{0.030}  &     \colorMyCellMorphFemaleInception{0.001}  &    \colorMyCellMorphFemaleInception{0.0} &  \colorMyCellMorphFemaleInception{0.0}\\
      \hline
      & \multicolumn{6}{c|}{$\text{FNMR}_x(\tau)$} \\
      
      \hline
  White - White & \colorMyCellMorphFemaleInception{0.0}  &   \colorMyCellMorphFemaleInception{0.004} &  \colorMyCellMorphFemaleInception{0.058}  & \colorMyCellMorphFemaleInception{0.152}  & \colorMyCellMorphFemaleInception{0.37} & \colorMyCellMorphFemaleInception{0.525} \\
  Black - Black & \colorMyCellMorphFemaleInception{0.0}  &   \colorMyCellMorphFemaleInception{0.0}   &  \colorMyCellMorphFemaleInception{0.005}  & \colorMyCellMorphFemaleInception{0.041}  & \colorMyCellMorphFemaleInception{0.098} & \colorMyCellMorphFemaleInception{0.196}\\

       \hline
        $FDR(\tau)$ & 0.7965 &  0.923  &  0.959  &  0.9425 &  0.8635  & 0.8355 \\
       \hline        
      
   \end{tabular}

  \label{tb:morph_female_inception}
  \end{center}  
\end{table}

\begin{table}[ht]
   \begin{center}

   \caption{MORPH-Female cohort - ArcFace-InsightFace: $\text{FNMR}(\tau)$, $\text{FMR}(\tau)$, and $\text{FDR}(\tau)$ per demographic (Demog.) in the test set. These figures of merit are fragmented by the race of the samples used for enrollment and the race of the samples used for probe (``(e-p)'' in the table.). In this example $\tau=\text{FMR}^{\text{dev}}_{x}$ where \textbf{dev} is the development set.}

    \begin{tabular}{|r||rrrrrr|}
      \hline
  
     $\tau=FMR_{10^{-x}}$ &  $10^{-1}$ &   $10^{-2}$ &   $10^{-3}$ &   $10^{-4}$ &   $10^{-5}$ & $10^{-6}$ \\
	
      \hline
      Demog (e-p) & \multicolumn{6}{c|}{$\text{FMR}_x(\tau)$} \\
      \hline
  White - White & \colorMorphFemFace{0.187} & \colorMorphFemFace{0.024}   & \colorMorphFemFace{0.002}   & \colorMorphFemFace{0.0}  & \colorMorphFemFace{0.0} & \colorMorphFemFace{0.0}\\
  White - Black & \colorMorphFemFace{0.088} & \colorMorphFemFace{0.003}  & \colorMorphFemFace{0.0}      & \colorMorphFemFace{0.0}  & \colorMorphFemFace{0.0} & \colorMorphFemFace{0.0}\\
  Black - White & \colorMorphFemFace{0.088} & \colorMorphFemFace{0.003}  & \colorMorphFemFace{0.0}      & \colorMorphFemFace{0.0}  & \colorMorphFemFace{0.0} & \colorMorphFemFace{0.0}\\
  Black - Black & \colorMorphFemFace{0.285} & \colorMorphFemFace{0.06}   & \colorMorphFemFace{0.01}     & \colorMorphFemFace{0.001} & \colorMorphFemFace{0.0} & \colorMorphFemFace{0.0}\\
      \hline
      & \multicolumn{6}{c|}{$\text{FNMR}_x(\tau)$} \\
      
      \hline
  White - White & \colorMorphFemFace{0.0}  &   \colorMorphFemFace{0.0}   &  \colorMorphFemFace{0.0}    & \colorMorphFemFace{0.0}  & \colorMorphFemFace{0.007}   & \colorMorphFemFace{0.011} \\
  Black - Black & \colorMorphFemFace{0.0}  &   \colorMorphFemFace{0.0}   &  \colorMorphFemFace{0.0}    & \colorMorphFemFace{0.0}  & \colorMorphFemFace{0.0} & \colorMorphFemFace{0.002}\\

       \hline
        $FDR(\tau)$ & 0.951  & 0.982  &  0.996  &  0.9995 &  0.9965 &  0.9955 \\
       \hline        
      
   \end{tabular}

  \label{tb:morph_female_insight}
  \end{center}  
\end{table}

\begin{table}[ht]
   \begin{center}

   \caption{MORPH-Female cohort - Resnet50: $\text{FNMR}(\tau)$, $\text{FMR}(\tau)$, and $\text{FDR}(\tau)$ per demographic (Demog.) in the test set. These figures of merit are fragmented by the race of the samples used for enrollment and the race of the samples used for probe (``(e-p)'' in the table.). In this example $\tau=\text{FMR}^{\text{dev}}_{x}$ where \textbf{dev} is the development set.}

    \begin{tabular}{|r||rrrrrr|}
      \hline
  
      $\tau=FMR_{10^{-x}}$ &  $10^{-1}$ &   $10^{-2}$ &   $10^{-3}$ &   $10^{-4}$ &   $10^{-5}$ & $10^{-6}$ \\
	
      \hline
      Demog (e-p) & \multicolumn{6}{c|}{$\text{FMR}_x(\tau)$} \\
      \hline
  White - White & \colorMorphFemGabor{0.193} & \colorMorphFemGabor{0.023} & \colorMorphFemGabor{0.003}   & \colorMorphFemGabor{0.0}  & \colorMorphFemGabor{0.0} & \colorMorphFemGabor{0.0}\\
  White - Black & \colorMorphFemGabor{0.064} & \colorMorphFemGabor{0.002} & \colorMorphFemGabor{0.0}   & \colorMorphFemGabor{0.0}  & \colorMorphFemGabor{0.0} & \colorMorphFemGabor{0.0}\\
  Black - White & \colorMorphFemGabor{0.063} & \colorMorphFemGabor{0.002} & \colorMorphFemGabor{0.0}   & \colorMorphFemGabor{0.0}  & \colorMorphFemGabor{0.0} & \colorMorphFemGabor{0.0}\\
  Black - Black & \colorMorphFemGabor{0.314} & \colorMorphFemGabor{0.058} & \colorMorphFemGabor{0.009} & \colorMorphFemGabor{0.001} & \colorMorphFemGabor{0.0} & \colorMorphFemGabor{0.0}\\
      \hline
      & \multicolumn{6}{c|}{$\text{FNMR}_x(\tau)$} \\
      
      \hline
  White - White & \colorMorphFemGabor{0.0}  &  \colorMorphFemGabor{0.014}   &  \colorMorphFemGabor{0.058} & \colorMorphFemGabor{0.167}  & \colorMorphFemGabor{0.304} & \colorMorphFemGabor{0.457} \\
  Black - Black & \colorMorphFemGabor{0.0}  &   \colorMorphFemGabor{0.005} &  \colorMorphFemGabor{0.019} & \colorMorphFemGabor{0.055}  & \colorMorphFemGabor{0.148} & \colorMorphFemGabor{0.227}\\

       \hline
        $FDR(\tau)$ & 0.9395 &  0.978  & 0.9775  &  0.9435  &  0.922  &  0.885 \\
       \hline        
      
   \end{tabular}

  \label{tb:morph_female_resnet}
  \end{center}  
\end{table}

\begin{table*}[ht]
   \begin{center}

   \caption{MORPH-Female cohort - COTS: $\text{FNMR}(\tau)$, $\text{FMR}(\tau)$, and $\text{FDR}(\tau)$ per demographic (Demog.) in the test set. These figures of merit are fragmented by the race of the samples used for enrollment and the race of the samples used for probe (``(e-p)'' in the table.). In this example $\tau=\text{FMR}^{\text{dev}}_{x}$ where \textbf{dev} is the development set.}

    \begin{tabular}{|r||rrrrrr|}
      \hline
  
      $\tau=FMR_{10^{-x}}$ &  $10^{-1}$ &   $10^{-2}$ &   $10^{-3}$ &   $10^{-4}$ &   $10^{-5}$ & $10^{-6}$ \\
	
      \hline
      Demog (e-p) & \multicolumn{6}{c|}{$\text{FMR}_x(\tau)$} \\
      \hline
  White - White & \colorMorphFemRankOne{0.373} & \colorMorphFemRankOne{0.082} & \colorMorphFemRankOne{0.013} & \colorMorphFemRankOne{0.003} & \colorMorphFemRankOne{0.0} & \colorMorphFemRankOne{0.0}\\
  White - Black & \colorMorphFemRankOne{0.133} & \colorMorphFemRankOne{0.009} & \colorMorphFemRankOne{0.0}   & \colorMorphFemRankOne{0.0}   & \colorMorphFemRankOne{0.0} & \colorMorphFemRankOne{0.0}\\
  Black - White & \colorMorphFemRankOne{0.141} & \colorMorphFemRankOne{0.012} & \colorMorphFemRankOne{0.001} & \colorMorphFemRankOne{0.0}   & \colorMorphFemRankOne{0.0} & \colorMorphFemRankOne{0.0}\\
  Black - Black & \colorMorphFemRankOne{0.239} & \colorMorphFemRankOne{0.044} & \colorMorphFemRankOne{0.006} & \colorMorphFemRankOne{0.001} & \colorMorphFemRankOne{0.0} & \colorMorphFemRankOne{0.0}\\
      \hline
      & \multicolumn{6}{c|}{$\text{FNMR}_x(\tau)$} \\

      \hline
  White - White & \colorMorphFemRankOne{0.0} &  \colorMorphFemRankOne{0.0} &  \colorMorphFemRankOne{0.0} & \colorMorphFemRankOne{0.023}  & \colorMorphFemRankOne{0.023} & \colorMorphFemRankOne{0.023} \\
  Black - Black & \colorMorphFemRankOne{0.0} &  \colorMorphFemRankOne{0.0} &  \colorMorphFemRankOne{0.0} & \colorMorphFemRankOne{0.0}  & \colorMorphFemRankOne{0.0} & \colorMorphFemRankOne{0.0}\\

       \hline
        $FDR(\tau)$ & 0.933  &  0.981  &   0.996  &  0.987 &  0.988 &  0.988 \\
       \hline        
      
   \end{tabular}

  \label{tb:morph_female_rank_one}
  \end{center}  
\end{table*}